\documentclass[10pt,twocolumn,letterpaper]{article}

\usepackage{cvpr,times,graphicx,amsmath,amssymb,subfig,multirow,makecell}
\usepackage[pagebackref=true,colorlinks=true,bookmarks=false,pagebackref=false]{hyperref}
\cvprfinalcopy 
\ifcvprfinal\usepackage{authblk}\fi

\newcommand{\app}{\raise.17ex\hbox{$\scriptstyle\sim$}}
\newcommand{\Caption}[1]{\caption{\small #1}\vspace{-2mm}}

\newcommand{\tss}[1]{\textsuperscript{#1}}

\newlength\savewidth\newcommand\shline{\noalign{\global\savewidth\arrayrulewidth
\global\arrayrulewidth 1pt}\hline\noalign{\global\arrayrulewidth\savewidth}}

\begin{document}

\title{Semantic Amodal Segmentation\vspace{-6mm}}
\author[1,2]{Yan Zhu}
\author[1]{Yuandong Tian}
\author[2]{Dimitris Mexatas}
\author[1]{Piotr Doll\'ar}
\affil[1]{Facebook AI Research (FAIR)}
\affil[2]{Department of Computer Science, Rutgers University\vspace{-3mm}}
\maketitle

\begin{abstract}

Common visual recognition tasks such as classification, object detection, and semantic segmentation are rapidly reaching maturity, and given the recent rate of progress, it is not unreasonable to conjecture that techniques for many of these problems will approach human levels of performance in the next few years. In this paper we look to the future: what is the next frontier in visual recognition?

We offer one possible answer to this question. We propose a detailed image annotation that captures information beyond the visible pixels and requires complex reasoning about full scene structure. Specifically, we create an \emph{amodal} segmentation of each image: the full extent of each region is marked, not just the visible pixels. Annotators outline and name all salient regions in the image and specify a partial depth order. The result is a rich scene structure, including visible and occluded portions of each region, figure-ground edge information, semantic labels, and object overlap.

We create two datasets for semantic amodal segmentation. First, we label 500 images in the BSDS dataset with multiple annotators per image, allowing us to study the statistics of human annotations. We show that the proposed full scene annotation is surprisingly consistent between annotators, including for regions and edges. Second, we annotate 5000 images from COCO. This larger dataset allows us to explore a number of algorithmic ideas for amodal segmentation and depth ordering. We introduce novel metrics for these tasks, and along with our strong baselines, define concrete new challenges for the community.

\end{abstract}

\begin{figure}\centering
\includegraphics[width=0.47\textwidth]{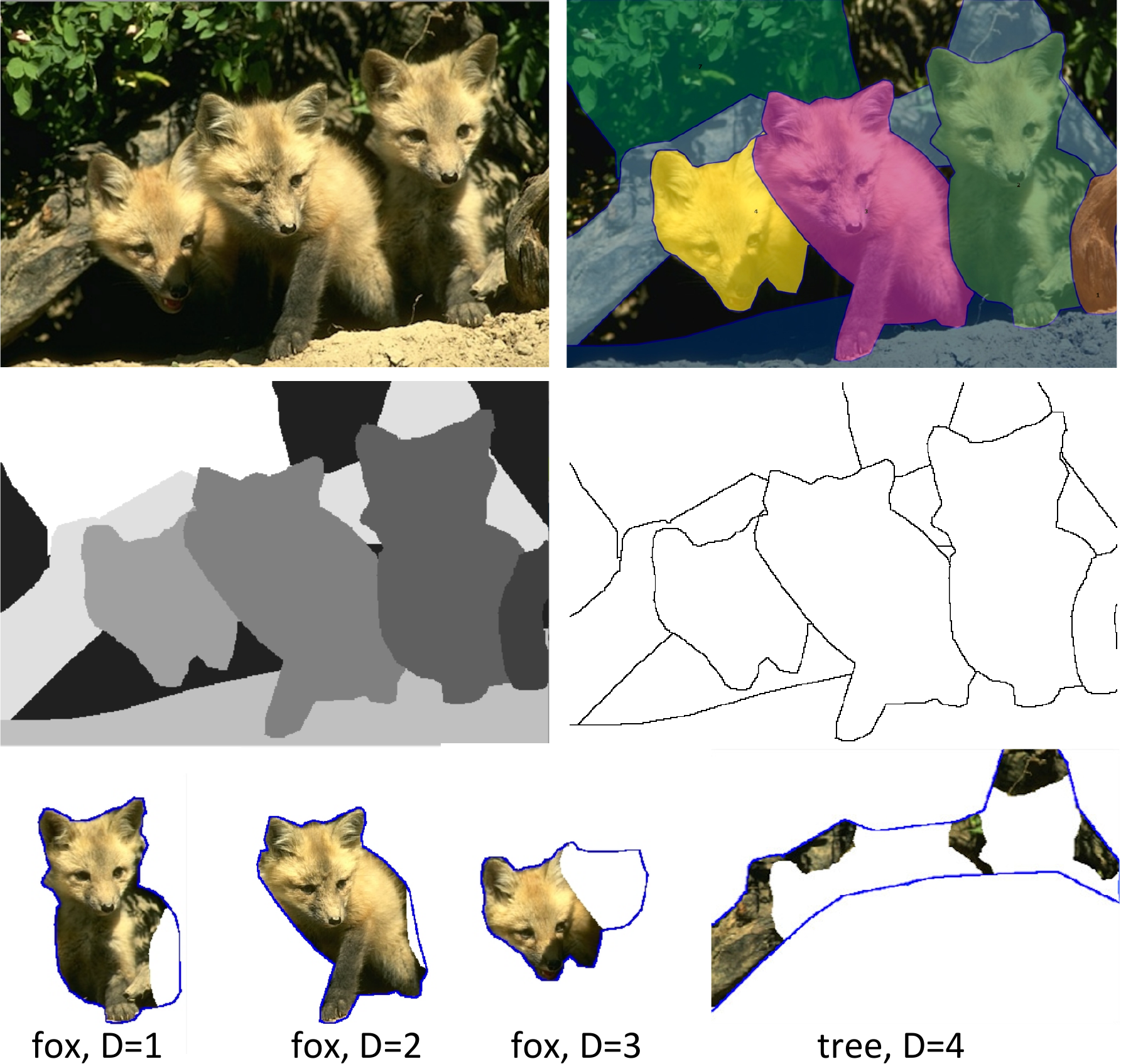}
\Caption{Example of \emph{Semantic Amodal Segmentation}. Given an image (top-left), annotators segment each region (top-right) and specify a partial depth order (middle-left). From this, visible edges can be obtained (middle-right) along with figure-ground assignment for each edge (not shown). All regions are annotated \emph{amodally}: the full extent of each region is marked, not just the visible pixels. Four annotated regions along with their semantic label and depth order are shown (bottom); note that both visible and occluded portions of each region are annotated.}
\label{fig:teaser}
\end{figure}

\section{Introduction}\label{sec:intro}

In recent years, visual recognition tasks such as image classification~\cite{AlexNet,He2016}, object detection~\cite{Felzenszwalb2009PAMI,SermanetICLR2013,Girshick2014rcnn,RenNIPS15faster}, edge detection~\cite{Arbelaez2011PAMI,dollar2015fast,xie2015holistically}, and semantic segmentation~\cite{Shotton2006ECCV,PinheiroICML2014,Long2015} have witnessed dramatic progress. This has been driven by the availability of large scale image datasets~\cite{Everingham10,imagenet_cvpr09,mscoco2015} coupled with a renaissance in deep learning techniques with massive model capacity~\cite{AlexNet,Simonyan15,GoogLeNet,He2016}. Given the pace of recent advances, one may conjecture that techniques for many of these tasks will rapidly approach human levels of performance. Indeed, preliminary evidence exists this is already the case for ImageNet classification~\cite{Karpathy2014blog}.

In this work we ask: what are the next set of challenges in visual recognition? What capabilities do we expect future visual recognition systems to possess?

We take our inspiration from the study of the human visual system. A remarkable property of human perception is the ease with which our visual system interpolates information not directly visible in an image~\cite{palmer1999vision}. A particularly prominent example of this, and one on which we focus, is \emph{amodal perception}: the phenomenon of perceiving the whole of a physical structure when only a portion of it is visible~\cite{kanizsa1979organization, palmer1999vision, wagemans2012century}. Humans can readily perceive partially occluded objects and guess at their true shape.

To encourage the study of machine vision systems with similar capabilities, we ask human subjects to annotate regions in images \emph{amodally}. Specifically, annotators are asked to mark the full extent of each region, not just the visible pixels. Annotators outline and name all salient regions in the image and specify a partial depth order. The result is a rich scene structure, including visible and occluded portions of each region, figure-ground edge information, semantic labels, and object overlap. See Figure~\ref{fig:teaser}.

An astute reader may ask: is amodal segmentation even a well-posed annotation task? More precisely, will multiple annotators agree on the annotation of a given image?

To study these questions, we asked multiple annotators to label all 500 images in the BSDS dataset~\cite{Arbelaez2011PAMI}. We designed the annotation task in a manner that encouraged annotators to consider object relationships and reason about scene geometry. This resulted in agreement between annotators that is surprisingly strong. In particular, our data has higher region and edge consistency than the original BSDS labels. Likewise, annotators tend to agree on the amodal completions. We report a thorough study of human performance on amodal segmentation using this data and also use it to train and evaluate state-of-the-art edge detectors.

In addition to the BSDS data, we annotate a second larger semantic amodal segmentation dataset using 5000 images from COCO~\cite{mscoco2015}. To achieve this scale, each image in COCO was annotated with just one expert annotator plus strict quality control. The dataset is divided into 2500/1250/1250 images for train/val/test, respectively. We introduce novel evaluation metrics for measuring amodal segment quality and pairwise depth-ordering of region segments. We do not currently use the semantic labels for evaluation as they come from an open vocabulary; nevertheless, we show that collecting these labels is key for obtaining high-quality amodal annotations. All train and val annotations along with evaluation code will be publicly released.

Finally, the larger collection of annotations on COCO allows us to train strong baselines for amodal segmentation and depth ordering. To perform amodal segmentation, we extend recent modal segmentation algorithms~\cite{pinheiro2015learning,pinheiro2016learning} to the amodal setting. We train two baselines: first, we train a deep net to directly predict amodal masks, second, motivated by~\cite{li2016amodal}, we train a model that takes a modal mask and attempts to expand it. Both variants achieve large gains over their modal counterparts, especially under heavy occlusion. We also experiment with deep nets for depth ordering and achieve accuracy over 80\%.

Our challenging new dataset, metrics, and strong baselines define concrete new challenges for the community and we hope that they will help spur novel research directions.

\begin{figure}\centering
\includegraphics[width=.45\textwidth]{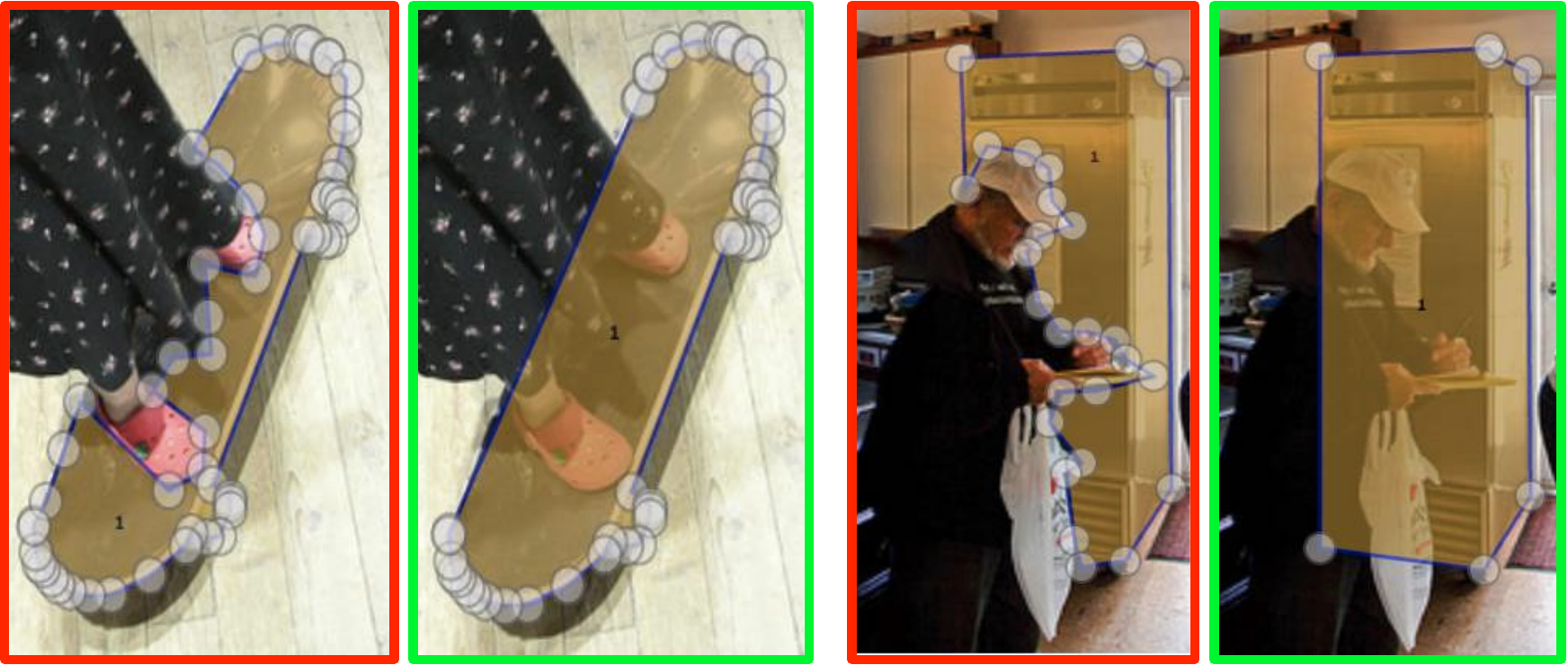}
\Caption{\emph{Amodal versus modal segmentation}: The left (red frame) of each image pair shows the modal segmentation of a region (visible pixels only) while the right (green frame) shows the amodal segmentation (visible and interpolated region). In this work we ask annotators to segment regions amodally. Note that the amodal segments have simpler shapes than the modal segments.}
\label{fig:examples_amodal}
\end{figure}

\subsection{Related Work}

Amodal perception~\cite{kanizsa1979organization} has been studied extensively in the psychophysics literature, for a review see~\cite{wagemans2012century,palmer1999vision}. However, amodal completion, along with many of the principles of perceptual grouping, are often demonstrated via simple illustrative examples such as the famous Kanizsa's triangle~\cite{kanizsa1979organization}. To our knowledge, there is no large scale dataset of amodally segmented natural images.

\emph{Modal segmentation}\footnote{In an abuse of terminology, we use \emph{modal segmentation} to refer to an annotation of only the visible portions of a region. This lets us easily differentiate it from \emph{amodal segmentation} (full region extent annotated).} datasets are more common. The most well known of these is the BSDS dataset~\cite{Arbelaez2011PAMI}, which has been used extensively for training and evaluating edge detection~\cite{Dollar2006CVPR, dollar2015fast, xie2015holistically} and segmentation algorithms~\cite{Arbelaez2011PAMI}. BSDS was later extended with figure-ground edge labels~\cite{fowlkes2007local}. A drawback of this annotation style is that it lacks clear guidelines, resulting in inconsistencies between annotators.

An alternative to unrestricted modal segmentation is \emph{semantic segmentation}~\cite{Shotton2006ECCV, liu2011nonparametric, Silberman2012ECCV}, where each image pixel is assigned a unique label from a fixed category set (\eg grass, sky, person). Such datasets have higher consistency than BSDS. However, the label set is typically small, individual objects are not delineated, and the annotations are modal. Notable exception are the StreetScenes dataset~\cite{bileschi2006streetscenes}, which contains a few categories which are labeled amodally, and PASCAL context~\cite{Mottaghi2014context}, which uses a large category set.

The closest dataset to ours is the hierarchical scenes dataset from Maire \etal~\cite{maire2013hierarchical}, which aims to captures occlusion, figure-ground ordering, and object-part relations. The dataset consists of incredibly rich and detailed annotations for 100 images. Our dataset shares some similarities but is easier to collect, allowing us to scale. Likewise, Visual Genome~\cite{krishna2016visualgenome} also provides rich annotations, including depth ordering, but does not include segmentation.

Compared to \emph{object detection} datasets~\cite{Everingham10, imagenet_cvpr09, mscoco2015}, our annotation is dense, amodal, and covers both objects and regions. Related datasets such as LabelMe~\cite{Russell2008IJCV} and Sun~\cite{xiao2010sun} also have objects annotated modally. Only for pedestrian detection~\cite{Dollar2011PAMI} are objects often annotated amodally (with both visible and amodal bounding boxes).

We note that our annotation scheme subsumes modal segmentation~\cite{Arbelaez2011PAMI}, edge detection~\cite{Arbelaez2011PAMI}, and figure-ground edge labeling~\cite{fowlkes2007local}. As our COCO annotations (5000 images) are an order of magnitude larger than BSDS (500 images)~\cite{Arbelaez2011PAMI}, the previous de-facto dataset for these tasks, we expect our data to be quite useful for these classic tasks.

Finally there has been some algorithmic work on amodal completion~\cite{guo2012beyond, Gupta2013CVPR, silberman2014contour, kar2015amodal}. Of particular interest, Ke \etal~\cite{li2016amodal} recently proposed a general approach for amodal segmentation that serves as the foundation for one of our baselines (see \S\ref{sec:evaluation}). Most existing recognition systems, however, operate on a per-patch or per-window basis, or with a limited receptive field, including for object detection~\cite{Felzenszwalb2009PAMI, SermanetICLR2013, Girshick2014rcnn}, edge detection~\cite{Dollar2006CVPR, dollar2015fast, xie2015holistically}, and semantic segmentation~\cite{Shotton2006ECCV, PinheiroICML2014, Long2015}. Our dataset will present challenges to such methods as amodal segmentation requires reasoning about object interactions.

\section{Dataset Annotation}\label{sec:annotation}

\begin{figure}\centering
\includegraphics[width=.48\textwidth]{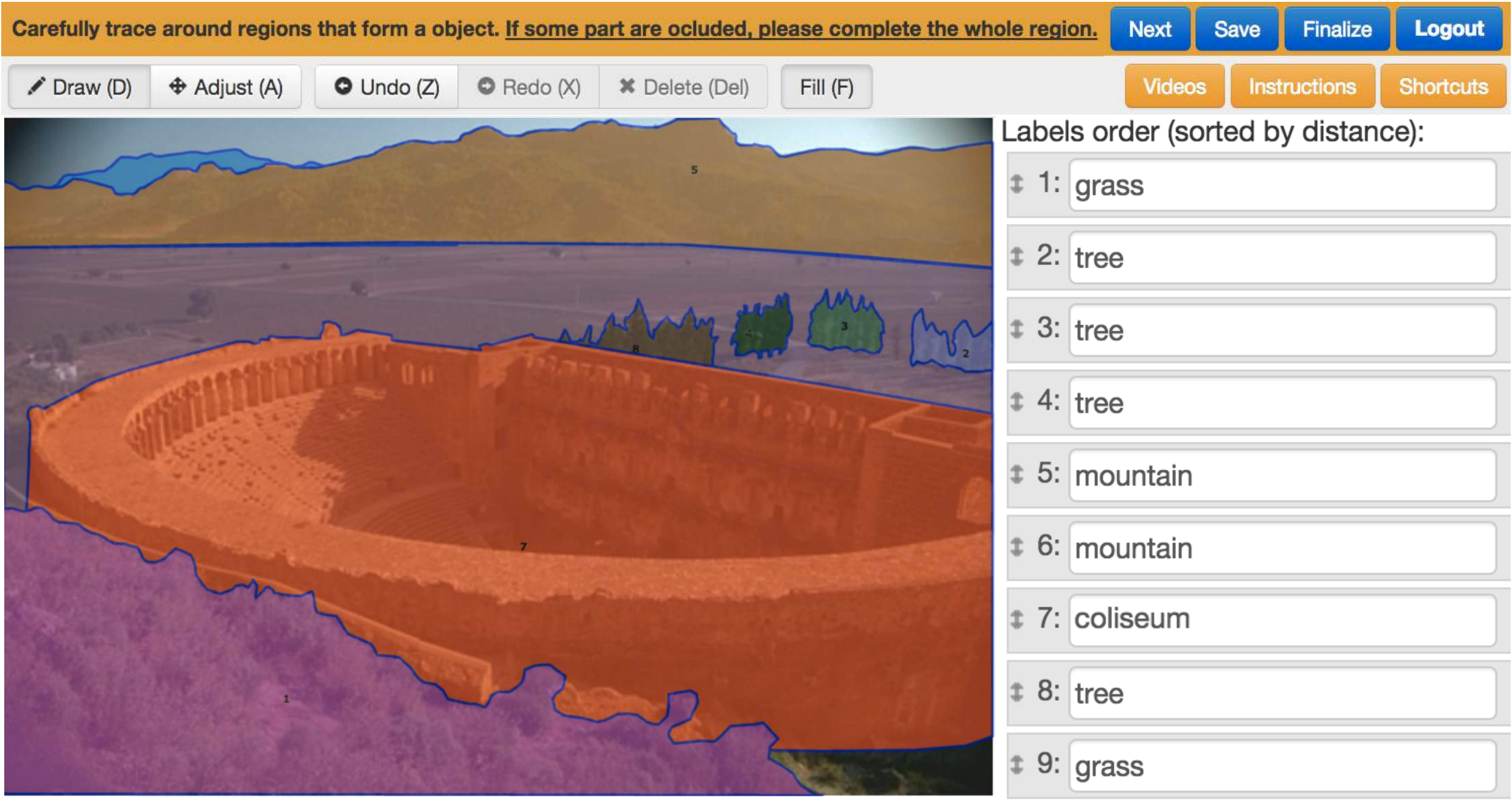}
\Caption{A screenshot of our annotation tool for semantic amodal segmentation (adopted from the Open Surfaces tool~\cite{bell2013opensurfaces}).}
\label{fig:screenshot}
\end{figure}

For our semantic amodal segmentation, we extend the Open Surfaces annotation tool from Bell \etal~\cite{bell2013opensurfaces}, see Figure~\ref{fig:screenshot}. The original tool allows for labeling multiple regions in an image by specifying a closed polygon for each; the same tool was also adopted for annotation of COCO~\cite{mscoco2015}. We extend the tool in a number of ways, including for region ordering, naming, and improved editing. For full details, including handling of corner cases, we refer readers to the supplementary. We will open-source the updated tool. 

\begin{figure}\centering
\includegraphics[width=.48\textwidth]{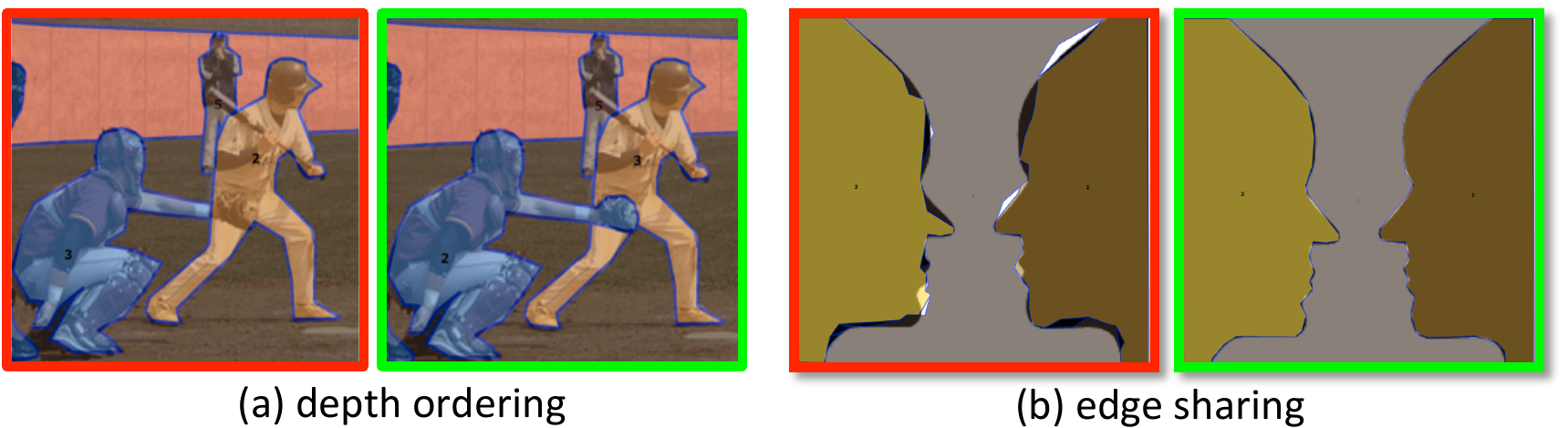}
\Caption{(a) We ask annotators to arrange region depth order. The right panel gives a correct depth order of the two people in the foreground while in the left panel the order is reversed. (b) Shared region edges must be marked to avoid duplicate edges. Unlike regular edges, shared edges do not have a figure-ground side.}
\label{fig:depth+sharing}
\end{figure}

We found four guidelines to be key for obtaining high-quality and consistent annotations: (1) only semantically meaningful regions should be annotated, (2) images should be annotated densely, (3) all regions should be ordered in depth, and (4) shared region boundaries should be marked. These guidelines encouraged annotators to consider object relationships and reason about scene geometry, and have proven to be effective in practice as we show in \S\ref{sec:consistency}.

\textit{(1) Semantic annotation:} Annotators are asked to name all annotated regions. Perceptually, the fact that a segment can be named implies that it has a well-defined prototype and corresponds to a semantically meaningful region. This criterion leads to a natural constraint on the granularity of the annotation: material boundaries and object parts (\ie interior edges) should not be annotated if they are not namable. Moreover, under this constraint, annotators are more likely to have a consistent prior on the occluded part of a region. In practice, we found that enforcing region naming led to more consistent and higher-quality amodal annotations.

\textit{(2) Dense annotation:} Annotators are asked to label an image densely, in particular all foreground object over a minimum size (600 pixels) should be labeled. Of particular importance is that if an annotated region is occluded, the occluder should also be annotated. When all foreground regions are annotated and a depth order specified, the visible and occluded portions of each annotated region are determined, as are the visible and hidden edges.

\textit{(3) Depth ordering:} Annotators are asked to specify the relative depth order of all regions, see Figure~\ref{fig:depth+sharing}a. In particular, for two overlapping regions, the occluder should precede the occludee. In ambiguous cases, the depth order is specified so that edges are correctly `rendered' (\eg, eyes go in front of the face). For non-overlapping regions any depth order is acceptable. Depth ordering encourages annotators to reason about scene geometry, including occlusion, and therefore improves the quality of amodal annotation.

\textit{(4) Edge sharing:} When one region occludes another, the figure-ground relation is clear, and an edge separating the regions belongs to the foreground region. However, when two regions are adjacent, an edge is shared and has no figure-ground side. We require annotators to explicitly mark shared edges, thus avoiding duplicate edges, see Figure \ref{fig:depth+sharing}b. As with the other criteria, this encourages annotators to reason about object interactions and scene geometry.

We refer readers to the supplementary material for additional details on the annotation tool and pipeline.

\section{Dataset Statistics}\label{sec:stats}

\begin{figure}
\centering\footnotesize
\subfloat[dataset summary statistics\label{table:stats}]{
\renewcommand\arraystretch{1}
\addtolength{\tabcolsep}{-5pt}
\begin{tabular}[b]{l|cc}
  & BSDS  & COCO\\\shline
  ann/image      & 5-7  & 1\\
  regions/ann     & 7.3  & 9.2\\
  points/region   & 64   & 46\\
  pixel coverage  & 84\% & 69\%\\
  occlusion rate  & 62\% & 61\%\\
  occ/region      & 21\% & 31\%\\
  time/polygon    & 68s  & 41s\\
  time/region     & 2m   & 2m\\
  time/ann        & 15m  & 18m\vspace{0mm}\\
\end{tabular}}
\subfloat[most common semantic labels\label{fig:cloud}]{
\includegraphics[width=.54\linewidth]{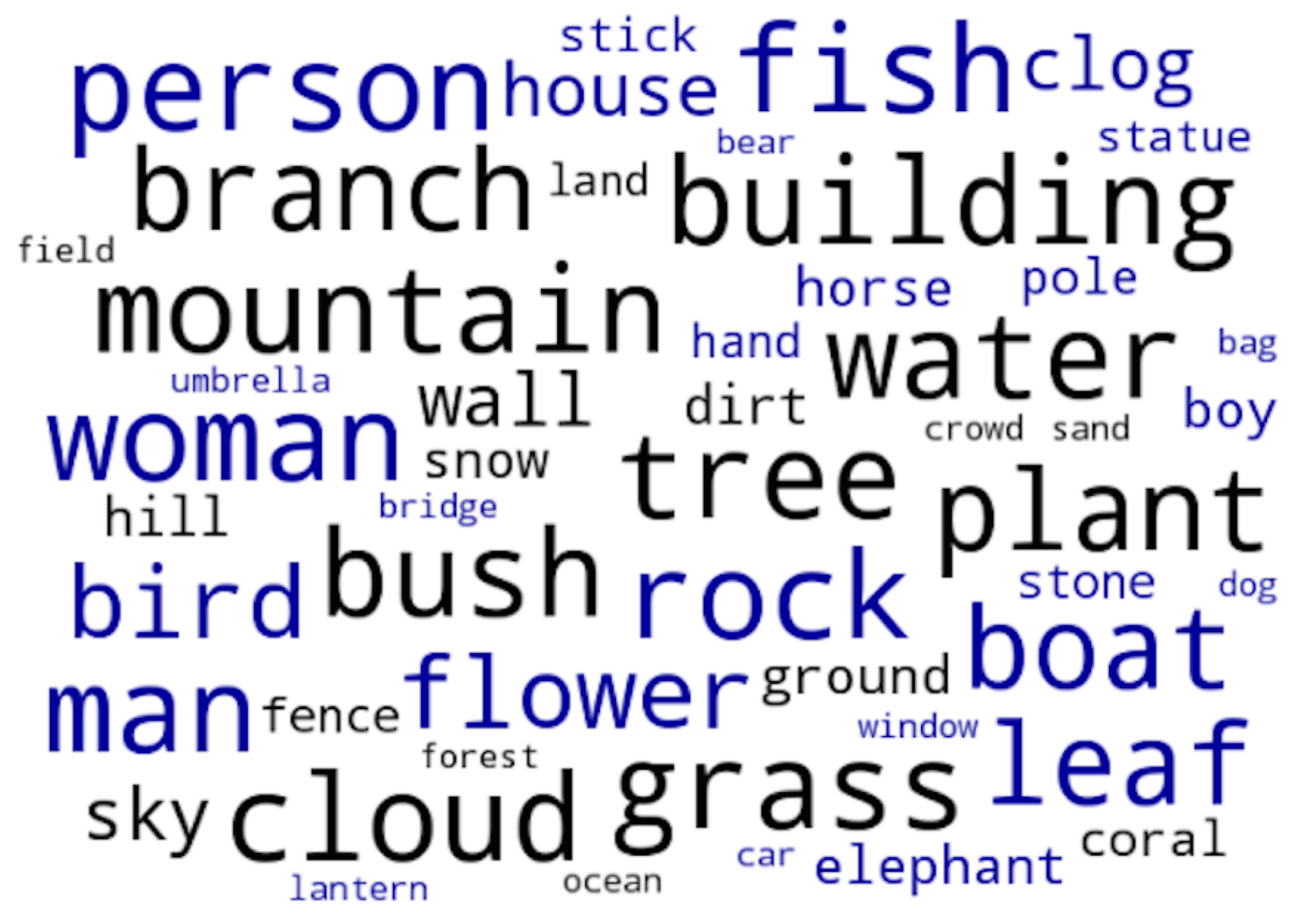}}
\Caption{(a) Dataset summary statistics on BSDS and COCO. COCO images are more cluttered, leading to some differences in statistics (\eg higher regions/ann and lower pixel coverage). (b) The top 50 semantic labels in our BSDS annotations. Roughly speaking, the blue words indicate `things' (person, fish, flower) while the black words indicate `stuff' (grass, cloud, water).}
\end{figure}

The analysis in this section is primarily based on the 500 images in the BSDS dataset~\cite{Arbelaez2011PAMI}, which has been used extensively for edge detection and modal segmentation. Annotating the same images amodally allows us to compare our proposed annotations to the original annotations. While all following analysis is based on these images, we note that the statistics of our annotations on COCO~\cite{mscoco2015} are similar (they differ slightly as COCO images are more cluttered).

Figure~\ref{table:stats} summarizes the statistics of our data. Each of the 500 BSDS images was annotated independently by 5 to 7 annotators. On average each image annotation consists of 7.3 labeled regions, and each region polygon consists of 64 points. About 84\% of image pixels are covered by at least one region polygon. Of all regions, $62\%$ are partially occluded and average occlusion is $21\%$.

Annotating a single region takes \app2 minutes. Of this, half the time is spent on the initial polygon and the rest on naming, depth ordering, and polygon refinement. Annotating an entire image takes \app15m, although this varies based on image complexity and annotator skill.

\emph{Semantic labels:} Figure \ref{fig:cloud} shows the top 50 semantic labels in our data with word size indicating region frequency. The labels give insight into the regions being labeled as well as the granularity of the annotation. Most labels correspond to basic level categories and refer to entire objects (not object parts). Using common terminology~\cite{adelson1991plenoptic,forsyth1996finding}, we explicitly classify the labels into two categories: `things' and `stuff', where a `thing' is an object with a canonical shape  (person, fish, flower) while `stuff' has a consistent visual appearance but can be of arbitrary spatial extent (grass, cloud, water). Both `thing' and `stuff' labels are prevalent in our data (stuff composes about a quarter of our regions).

\emph{Shape complexity:} One important property of amodal segments is that they tend to have a relatively simple shape compared to modal segments that is independent of scene geometry and occlusion patterns (see Figure \ref{fig:examples_amodal}). We verify this observation with the following two statistics, shape \emph{convexity} and \emph{simplicity}, defined on a segment $S$:

\vspace{-2mm}{\small\begin{eqnarray}
  convexity(S) &=& \frac{Area(S))}{Area(ConvexHull(S))} \\
  simplicity(S) &=& \frac{\sqrt{4\pi*Area(S)}}{Perimeter(S)}
\end{eqnarray}}\vspace{-2mm}

A segment with a large convexity and simplicity value means it is simple (and both metrics achieve their maximum value of $1.0$ for a circle). Table \ref{table:shape} shows that amodal regions are indeed simpler than modal ones, which verifies our hypothesis. Due to their simplicity, amodal regions can actually be more efficient to label than modal regions.

We also compare to the original (modal) BSDS annotations (first column of Table \ref{table:shape}). Interestingly, the original BSDS annotations are even simpler than our modal annotations. Qualitatively it appears that the original annotators had a bias for simpler shapes and smoother boundaries.

\emph{Edge density:} The last row of Table \ref{table:shape} shows that our dataset has fewer visible edges marked than the original BSDS annotation (edge density is the percentage of image pixels that are edge pixels). This is necessarily the case as material boundaries and object parts (i.e.~interior edges) are not annotated in our data. Note that in \S\ref{sec:consistency} we demonstrate that although our edge maps are slightly less dense, they can be used to effectively train state-of-the-art edge detectors.

\begin{figure*}\centering
\subfloat[detailed occlusion statistics\label{fig:histOcclusion}]{
  \includegraphics[width=0.45\textwidth]{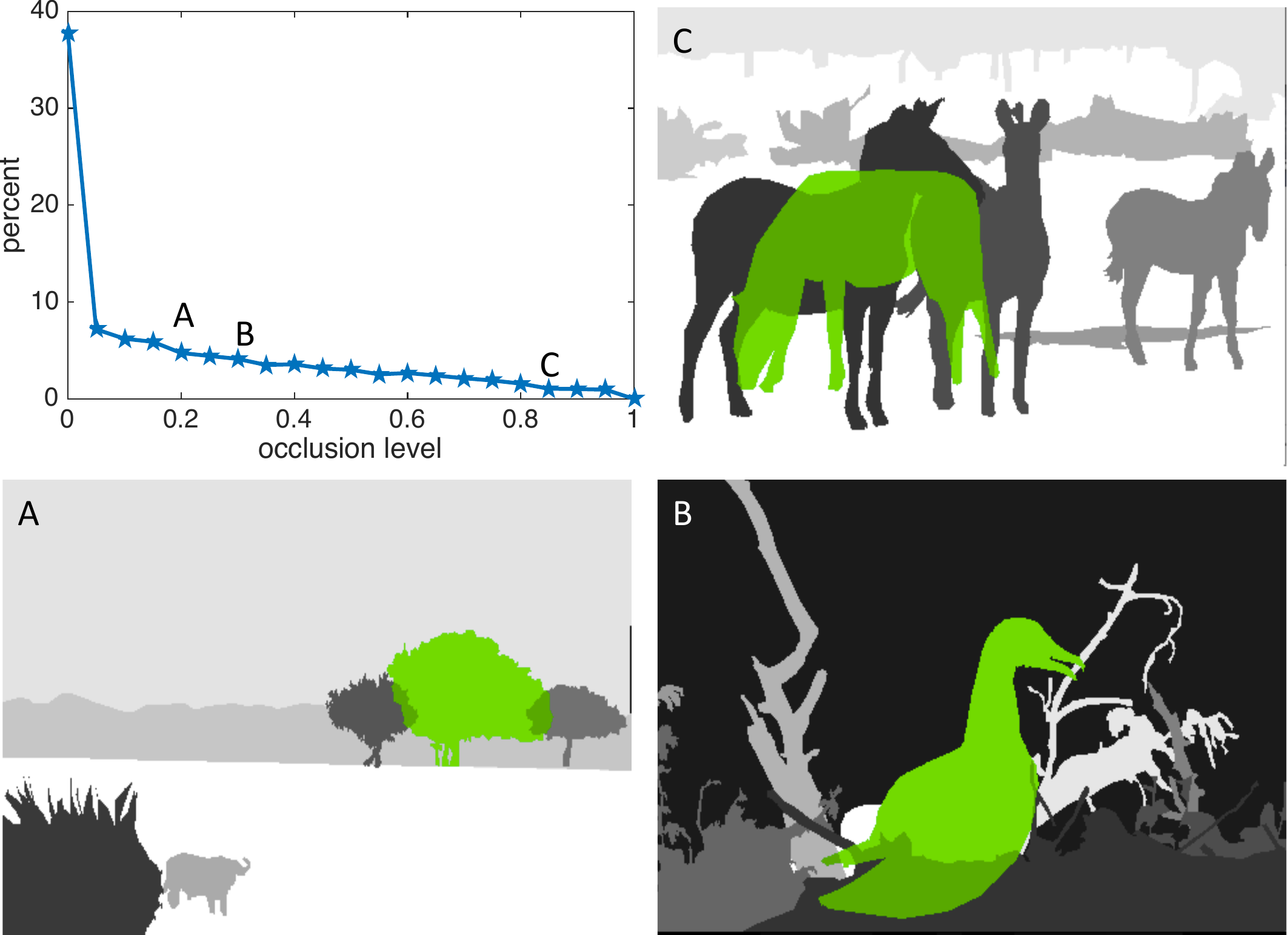}}\hspace{4mm}
\subfloat[number of connected components per annotation\label{fig:histCCNum}]{
  \includegraphics[width=0.45\textwidth]{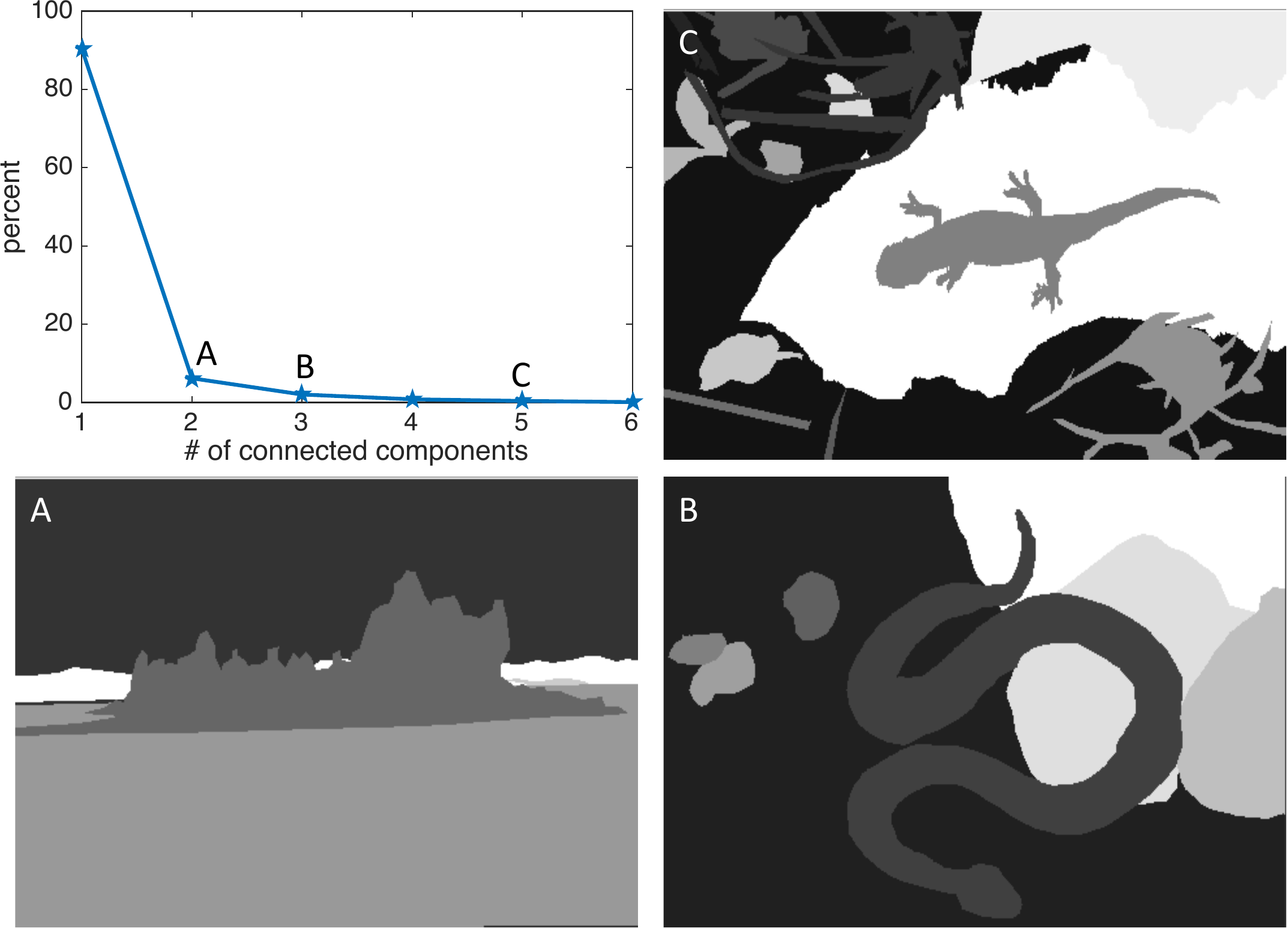}}\\
\subfloat[connected components size\label{fig:histCCsize}]{
  \includegraphics[width=0.45\textwidth]{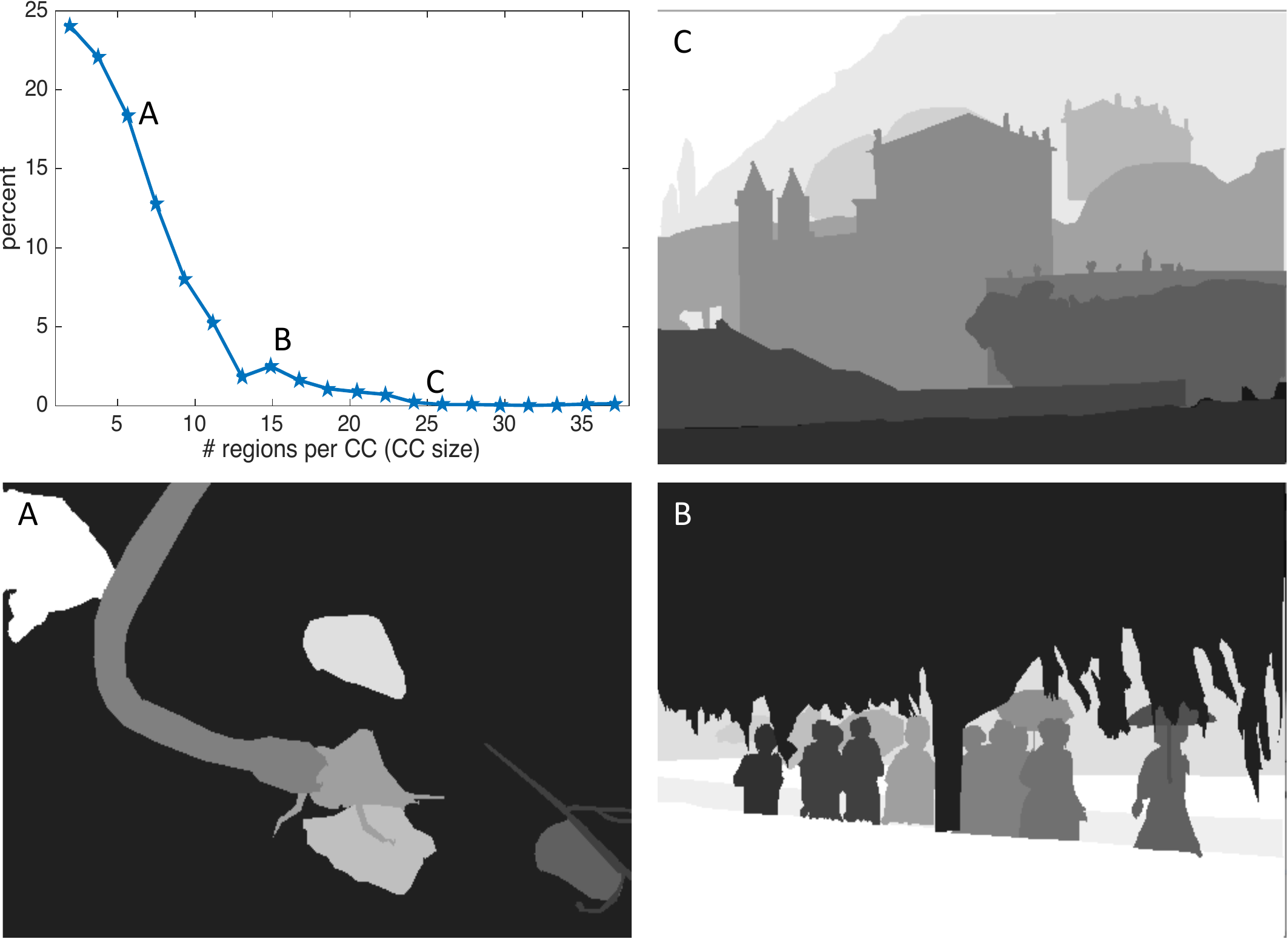}}\hspace{4mm}
\subfloat[number of depth layers per connected component\label{fig:histLayer}]{
  \includegraphics[width=0.45\textwidth]{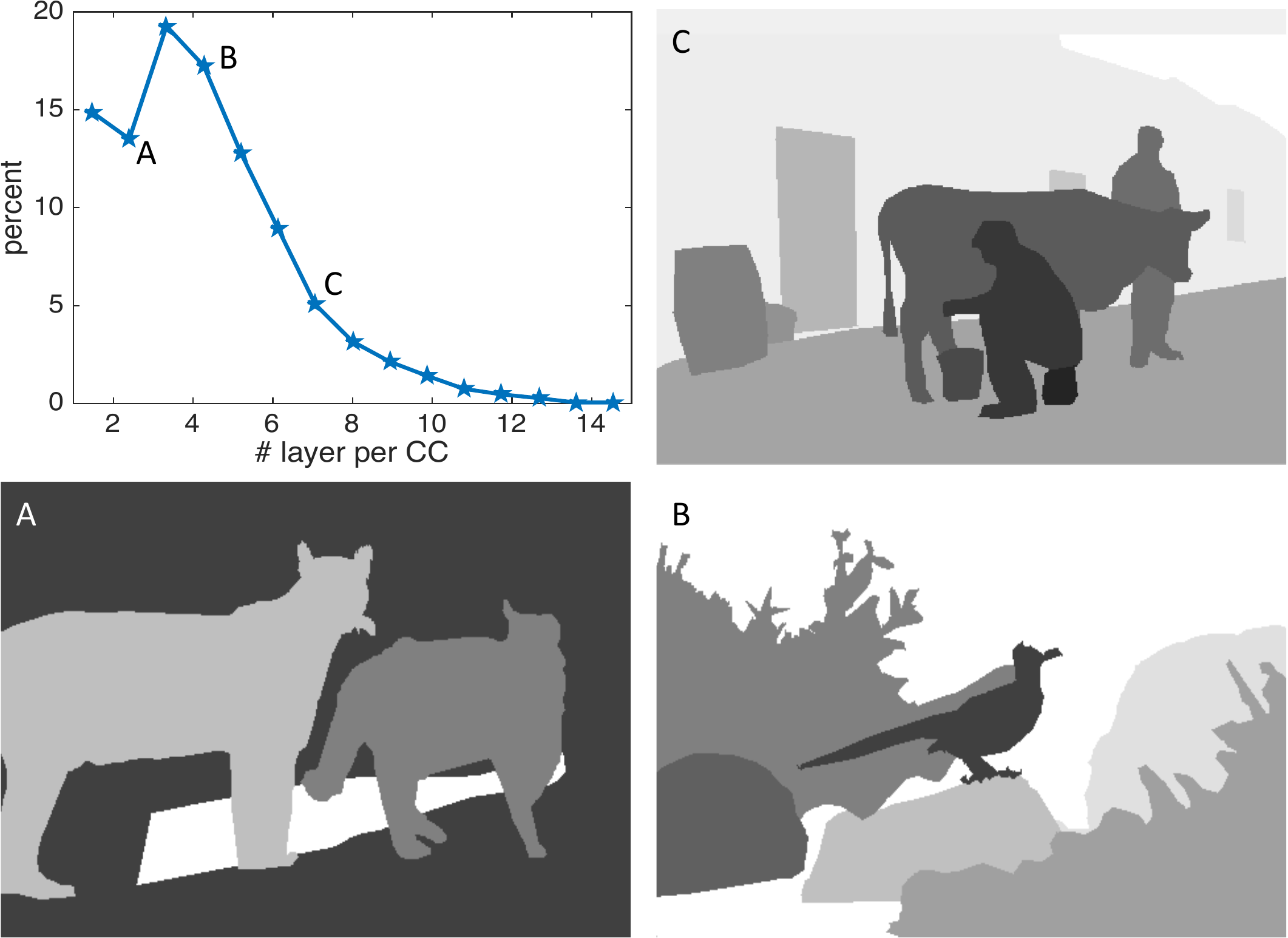}}
\Caption{Detailed dataset statistics. See text for details.}
\label{fig:Hist}
\end{figure*}

\emph{Occlusion:} Figure \ref{fig:histOcclusion} shows a histogram of occlusion level (defined as the fraction of region area that is occluded). Most regions are slightly occluded, while a small portion of regions are heavily occluded. We additionally display 3 occluded examples at different occlusion levels.

\begin{table}
\centering\small\renewcommand\arraystretch{1}
\renewcommand{\tabcolsep}{2mm}
\begin{tabular}{c|ccc|cc}
 & \multicolumn{3}{c|}{\textbf{BSDS}} & \multicolumn{2}{c}{\textbf{COCO}}\\
 & original & modal & amodal & modal & amodal\\
\shline
simplicity & .801   & .718   & .834   & .746   & .856\\
convexity  & .664   & .616   & .643   & .658   &.685\\
density    & 1.80\% & 1.57\% & 1.97\% & 1.71\% & 2.10\%\\
\end{tabular}
\Caption{Comparison of shape and edge statistics between modal and amodal segments on BSDS and COCO. Amodal segments tend to have a relatively simpler shape that is independent of scene geometry and occlusion patterns (see also Figure \ref{fig:examples_amodal}). Interestingly, the original BSDS annotations (first column) are even simpler than our modal annotations. Finally the last row reports edge density.}
\label{table:shape}
\end{table}

\emph{Scene complexity:} With the help of depth ordering, we can represent regions using a Directed Acyclic Graph (DAG). Specifically, we draw a directed edge from region $R_1$ to region $R_2$ if $R_1$ spatially overlaps $R_2$ and $R_1$ precedes $R_2$ in depth ordering. Given the DAG corresponding to an image annotation, a few quantities can be analyzed.

First, Figure \ref{fig:histCCNum} shows the number of connected components (CC) per DAG. Most annotations have only one CC, as shown in example A. If regions are scattered and disconnected an image will have more CC's, as in B and C.

The size of a CC measures how many regions are mutually overlapped, which in turns gives an implicit measure of scene complexity. Figure \ref{fig:histCCsize} shows a number of examples. More complex scenes (examples B and C) have large CC's.

Finally, the longest directed path of any CC in a DAG characterizes the minimum number of depth layers required to properly order all regions in the DAG. Note that the number of depth layers is often smaller than the size of a CC: e.g.~a large CC with numerous non-overlapping foreground objects and a single common background only requires two depth layers. Figure \ref{fig:histLayer} shows the distribution of number of depth layers needed per CC. Most components require only a few depth layers although some are far more complex.

\begin{figure}\centering
\includegraphics[width=0.33\textwidth]{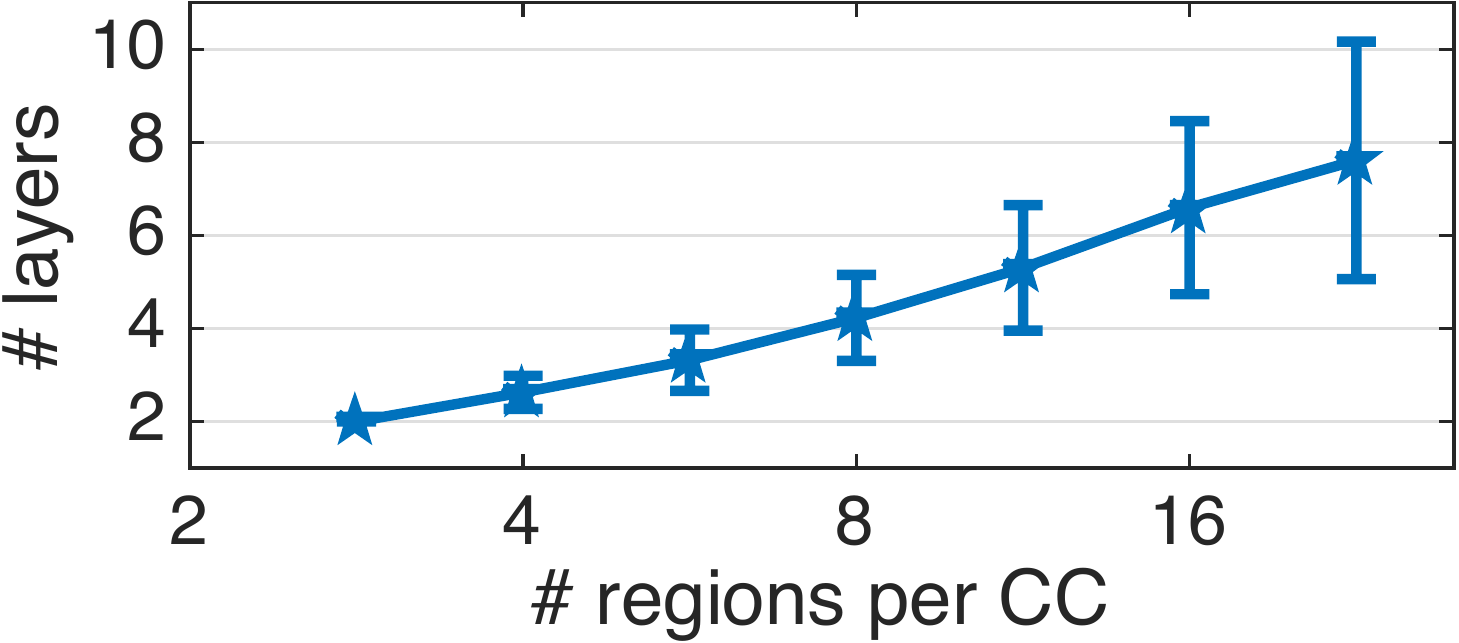}
\Caption{The minimum number of depth layers necessary to represent a connected component (CC). See text for details.}
\label{fig:layer}
\end{figure}

Figure \ref{fig:layer} further investigates the correlation between CC size and the minimum number of depth layers necessary to order all regions. We observe that the number of depth layers necessary appears to grow logarithmically with CC size.

\section{Dataset Consistency}\label{sec:consistency}

We next aim to show that semantic amodal segmentation is a well-posed annotation task. Specifically, we show that agreement between independent annotators is high. Consistency is a key property of any human-labeled dataset as it enables machine vision systems to learn a well defined concept. In the next two sub-sections we analyze our dataset's region and edge consistency on BSDS. As a baseline, we compare to the original (modal) BSDS annotations.

\begin{figure}\centering
\includegraphics[width=.23\textwidth]{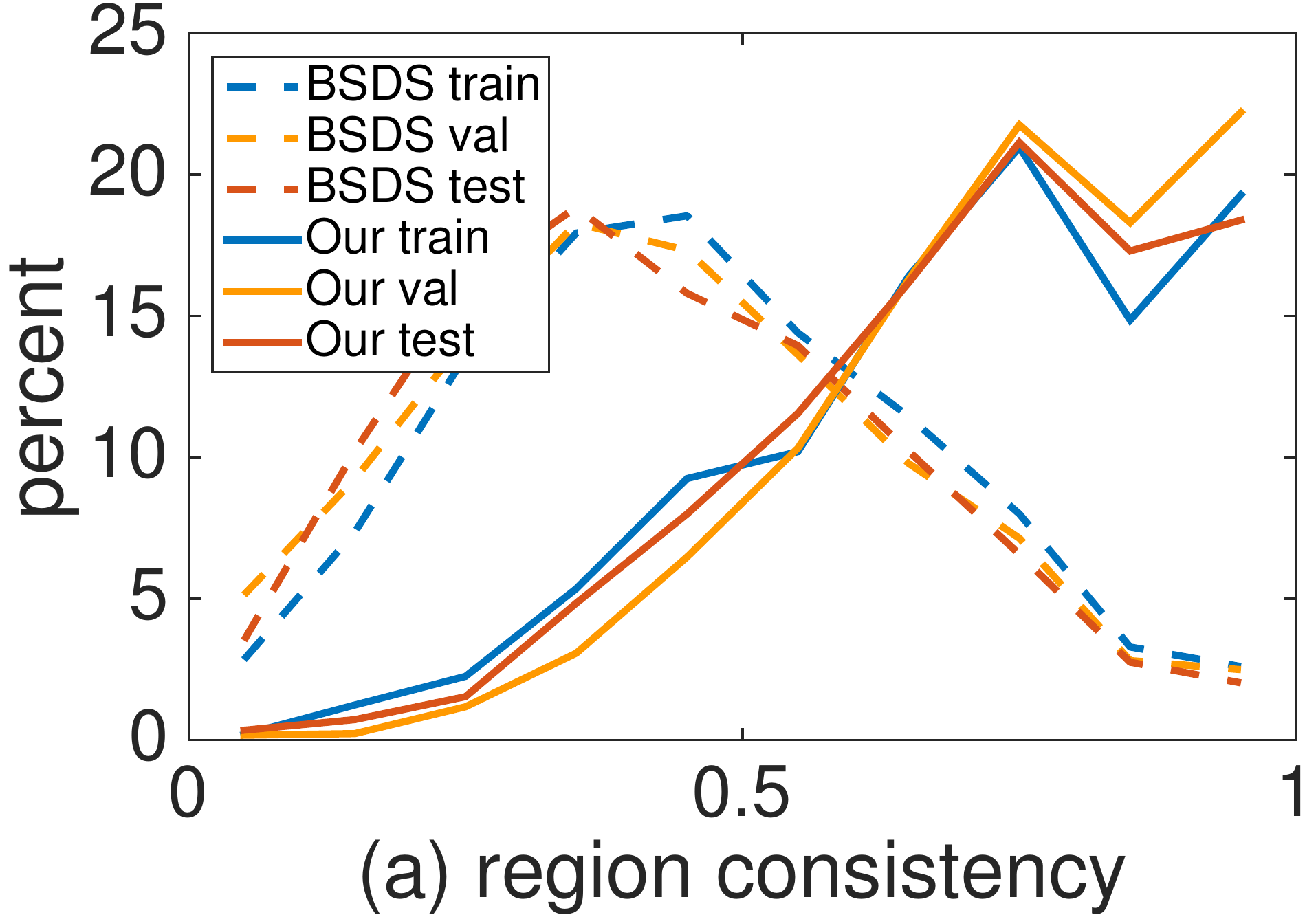}
\includegraphics[width=.23\textwidth]{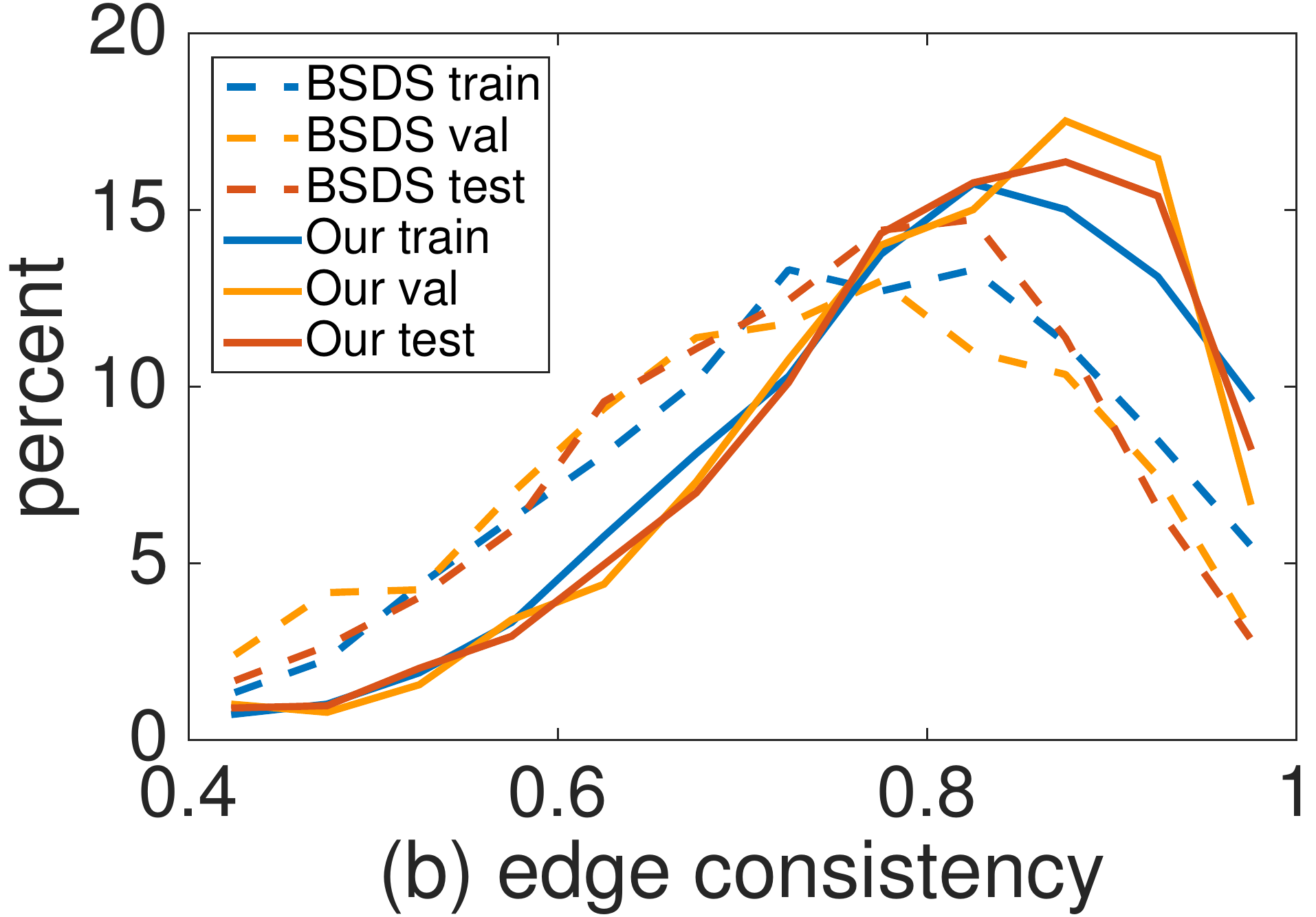}
\Caption{(a) Histogram of pairwise \emph{region consistency} scores for the original \emph{modal} BSDS annotations and our \emph{amodal} regions. (b) Histogram of pairwise \emph{edge consistency} scores for visible edges.}
\label{fig:consistency}
\end{figure}

\subsection{Region Consistency}

To measure region consistency, we use Intersection over Union (IoU) to match regions. The IoU between two segments is the area of their intersection divided by the area of their union. We threshold IoU at 0.5 and use bipartite matching to match two sets of regions. We set each annotation as the ground truth in turn, and for every other annotator we compute precision (P) and recall (R) and summarize the result via the $F$ measure: $F=2PR/(P+R)$. For $n$ annotators this yields $n(n-1)$ $F$ scores per image.

In Figure \ref{fig:consistency}a we display a histogram of $F$ scores for both the original BSDS \emph{modal} annotations from~\cite{Arbelaez2011PAMI} and the \emph{amodal} annotations in our proposed dataset across each split of the dataset. The region consistency of our amodal regions is substantially higher than the consistency of the original modal regions: median of 0.723 versus 0.425. This is in spite of the fact that our amodal regions include both the visible and occluded portions of each region. We note that the modal region consistency of our annotations is 0.756, slightly higher than for amodal regions, as expected.

A number of factors contribute to the consistency of our regions. Most importantly, we gave more focused instructions to the annotators; specifically, we asked annotators to label only semantically meaningful regions and to label all foreground objects, see \S\ref{sec:annotation}. Thus there was less inherent ambiguity in the task. Moreover, in modal segmentation, annotation level of detail substantially impacts region agreement.

Figure \ref{fig:qualitative} shows qualitative examples of annotator agreement on individual regions for both visible and occluded portions of a region. Naturally, annotations are most consistent for regions with simple shapes and little occlusions. On the other hand, when the object is highly articulated and/or severely occluded, annotators tend to disagree more.

\newcommand{\incg}[2]{
  \includegraphics[width=#1\linewidth]{figures/qualitative/#2-b.jpg}\hspace{.1mm}
  \includegraphics[width=#1\linewidth]{figures/qualitative/#2-a.jpg}}
\begin{figure*}
\centering\renewcommand\arraystretch{.25}
\renewcommand{\tabcolsep}{2mm}
\begin{tabular}{c@{\hskip -2mm}cccc}
  \multicolumn{4}{c}{\includegraphics[scale=0.5]{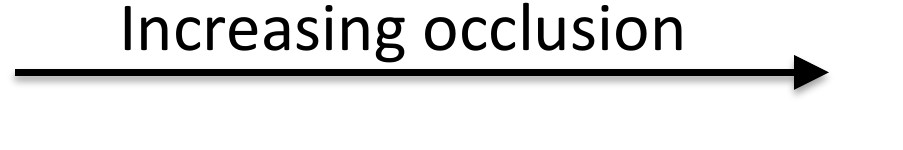}\vspace{-2.5mm}} \\
  \multirow{4}{*}{\includegraphics[scale=0.55]{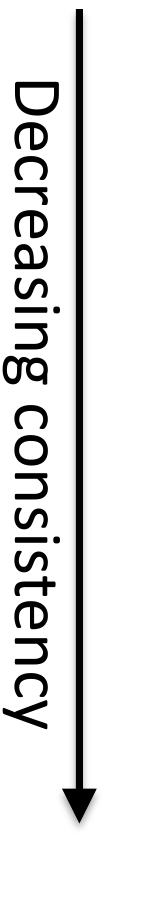}}
  & \incg{.10}{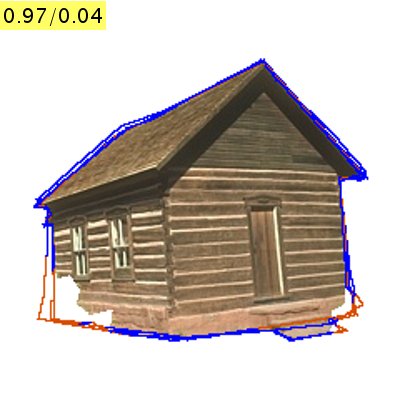} & \incg{.10}{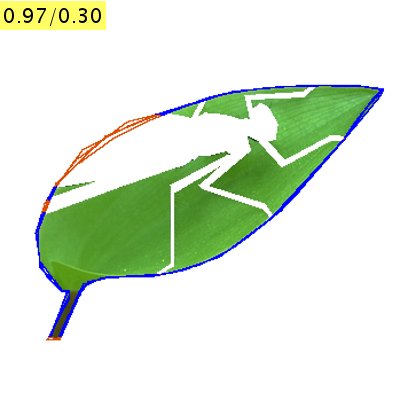} & \incg{.10}{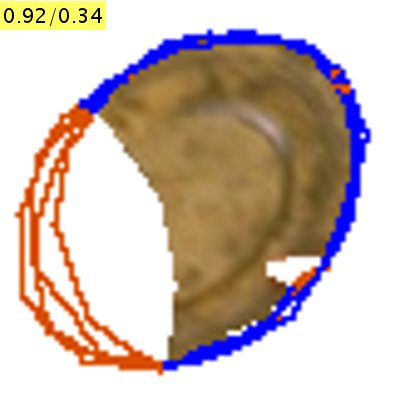} & \incg{.10}{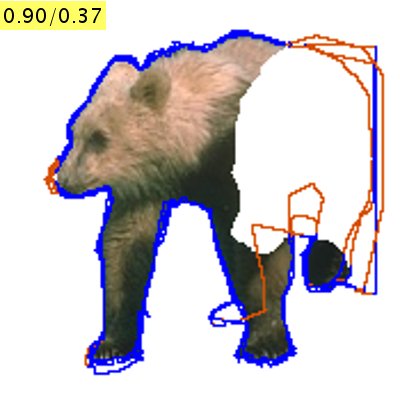}\\
  & \incg{.10}{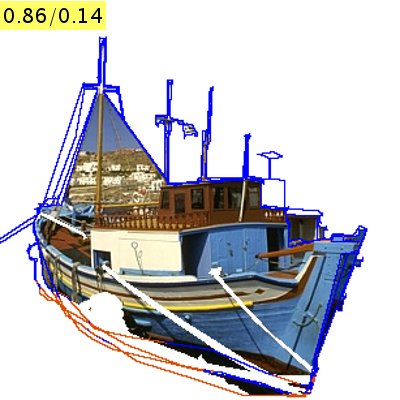}   & \incg{.10}{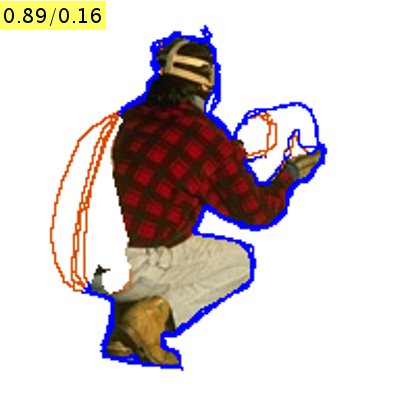}    & \incg{.10}{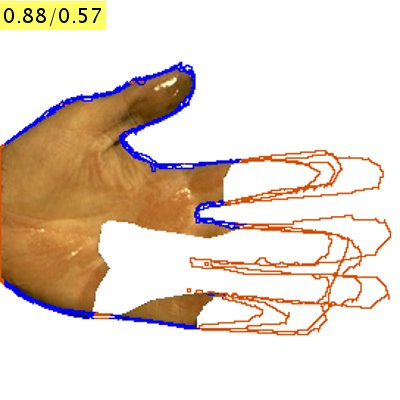}   & \incg{.10}{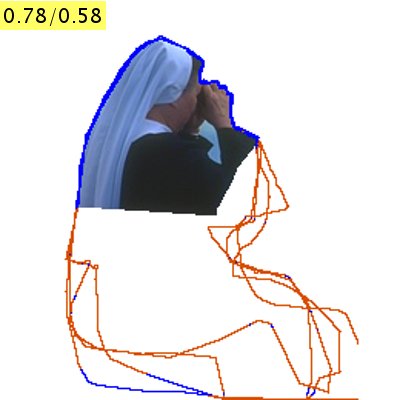}\\
  & \incg{.045}{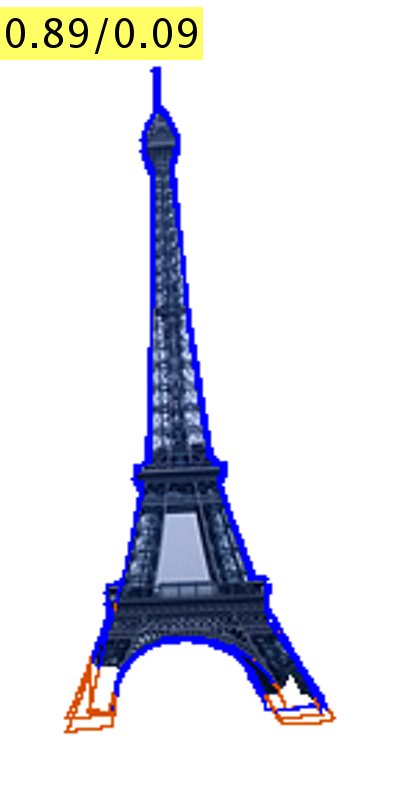}   \incg{.045}{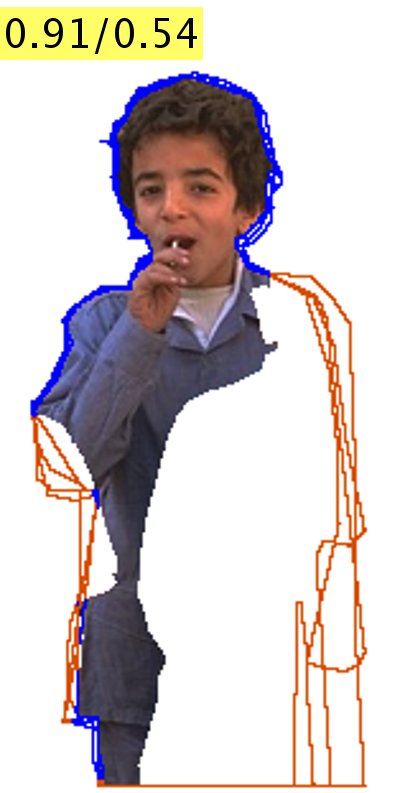}   & \incg{.045}{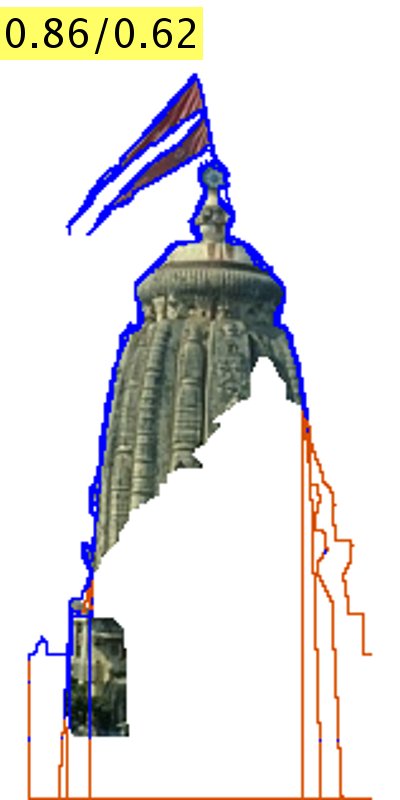}  \incg{.045}{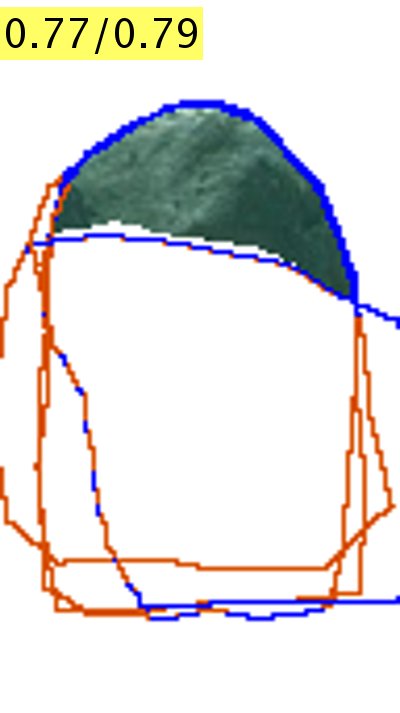}
  & \incg{.045}{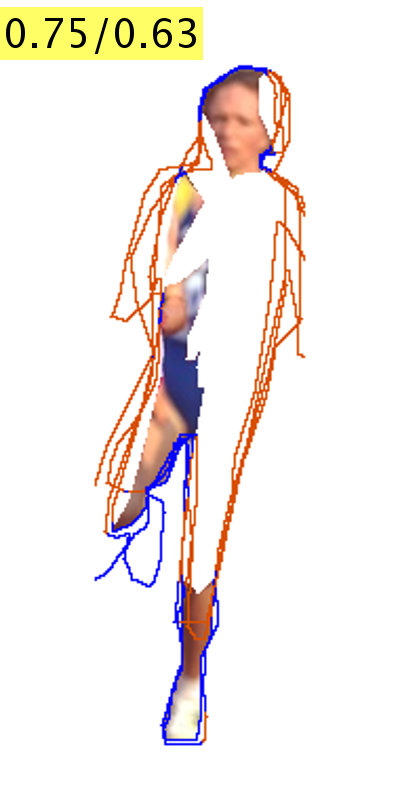}   \incg{.045}{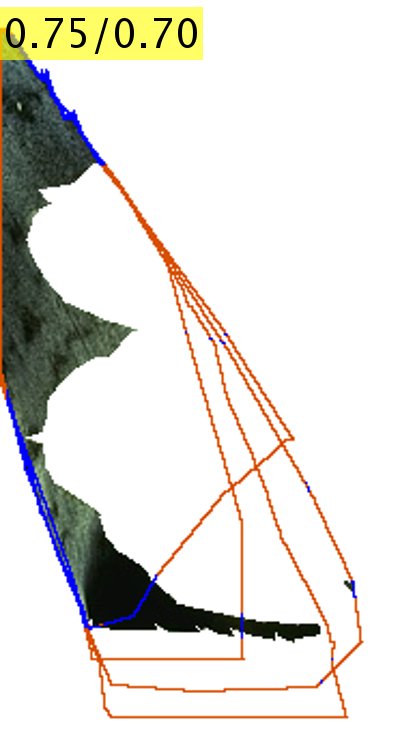} & \incg{.045}{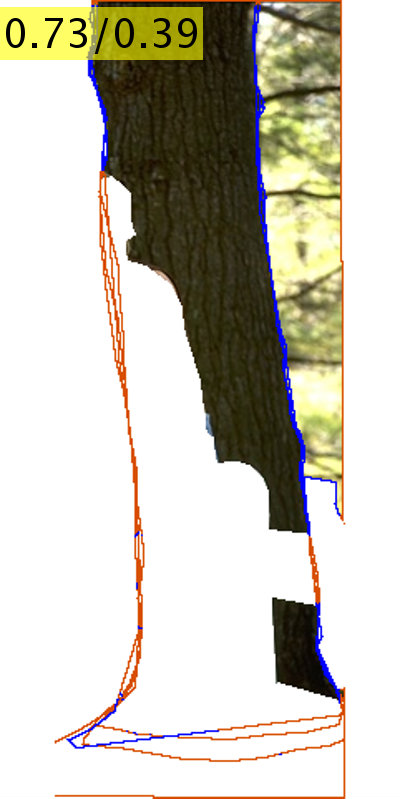}  \incg{.045}{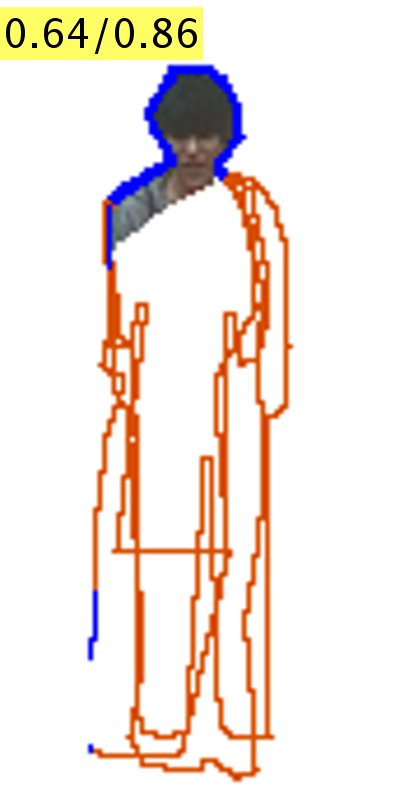}\\
  & \incg{.10}{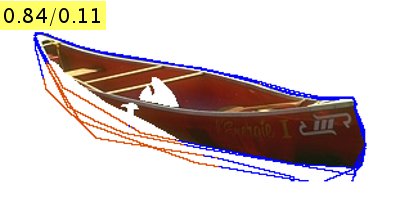}   & \incg{.10}{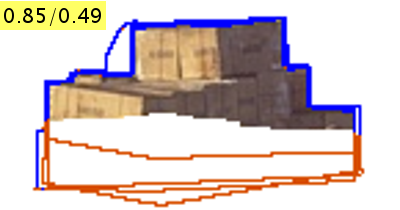}  & \incg{.10}{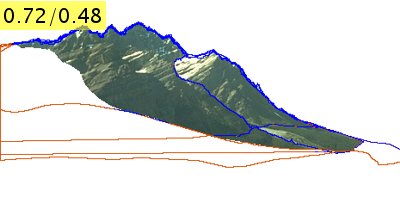}     & \incg{.10}{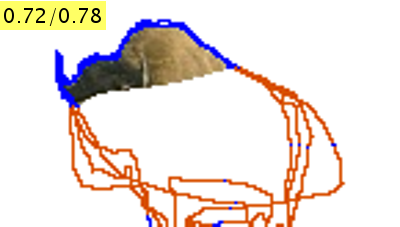}\\
  & \incg{.10}{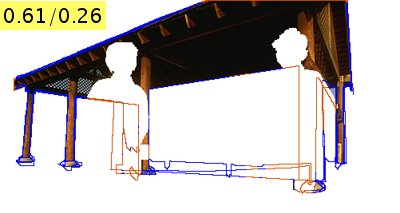} & \incg{.10}{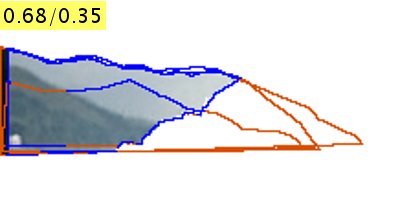} & \incg{.10}{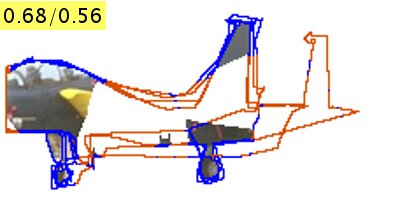}   & \incg{.10}{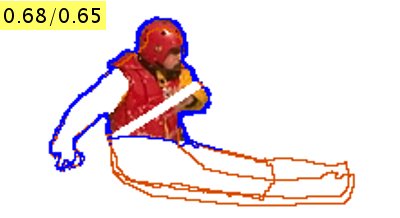}\\
  & \incg{.10}{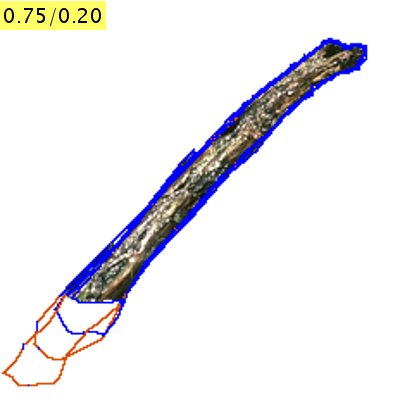}  & \incg{.10}{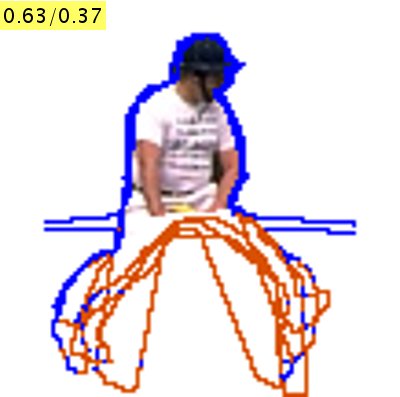}   & \incg{.10}{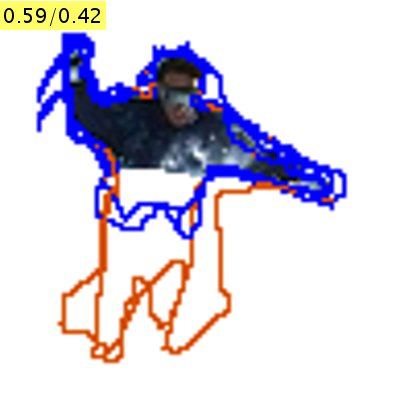}     & \incg{.10}{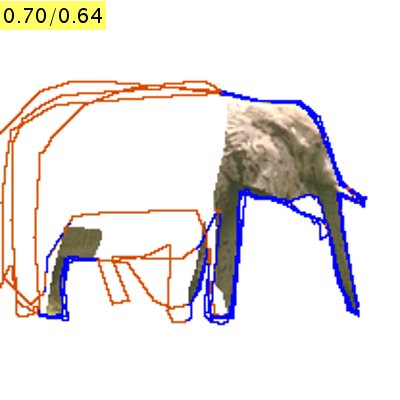}\\
\end{tabular}\vspace{-2mm}
\Caption{Visualizations of amodal region consistency. The blue edges are the visible edges, while the red edges are the occluded edges. Ground truth is determined by a single randomly chosen annotator. The region consistency score (average IoU score) and the occlusion rate are displayed. Examples are roughly sorted by decreasing consistency vertically and increasing occlusion horizontally.}
\label{fig:qualitative}\vspace{-2mm}
\end{figure*}

\subsection{Edge Consistency}\label{sec:edges}

\begin{table}
\centering\small\renewcommand\arraystretch{1}
\renewcommand{\tabcolsep}{2mm}
\begin{tabular}{c|ccc|ccc}
 &\multicolumn{3}{c|}{\textbf{SE}~\cite{dollar2015fast}} & \multicolumn{3}{c}{\textbf{HED}~\cite{xie2015holistically}}\\
  train / test & ODS & AP & R50 & ODS & AP & R50 \\
\shline
bsds / bsds & .744 & .795 & .921 & \textbf{.787} & .790 & .855\\
ours / bsds & \textbf{.747} & \textbf{.802} & \textbf{.923} & .775 & \textbf{.793} & \textbf{.868}\\
bsds / ours & .619 & .603 & .761 & .657 & \textbf{.578} & .697\\
ours / ours & \textbf{.630} & \textbf{.630} & \textbf{.785} & \textbf{.694} & .572 & \textbf{.752}\\
\end{tabular}
\Caption{Cross-dataset performance of two state-of-the-art edge detectors. For SE, training on our dataset improves performance even when testing on the original BSDS edges. For HED, using the same train/test combination maximizes performance. These results indicate that our dataset is valid for edge detection.}
\label{table:detectors}
\end{table}

Given the amodal annotations and depth ordering, along with the constraint that all foreground regions are annotated, we can compute the set of visible image edges. We next verify the quality of the obtained edge maps.

First, to measure edge consistency among annotators, we compute the F score between each pair of annotations, for details see~\cite{Arbelaez2011PAMI}. Figure \ref{fig:consistency}b shows the distribution of the boundary consistency scores. The edges in our amodal dataset are more consistent than edges in the original BSDS annotations (median consistency of 0.795 versus 0.728).

While our edges are more consistent, the edges are also less dense (see Table \ref{table:shape}). To evaluate the efficacy of using our data for edge detection, we test two popular state-of-the-art edge detectors: structured edges (SE)~\cite{dollar2015fast} and the holistically-nested edge detector (HED)~\cite{xie2015holistically}. Results for cross-dataset generalization are shown in Table \ref{table:detectors}. For SE, training on our dataset improves performance even when testing on the original BSDS edges. For HED, using the same train/test combination maximizes performance by a slight margin. These results indicate that our dataset is valid for edge detection. Note, however, that our test set is substantially harder as only semantic boundaries are annotated.

Finally, we measure human performance. As in \cite{Arbelaez2011PAMI}, we take one annotation as the detection and the union of the others as ground truth (note that this differs from the 1-vs-1 methodology used for Figure \ref{fig:consistency}b). On the original BSDS test set, precision/recall/F-Score are .92/.73/.81. Human performance is much higher on our test set, the scores are .98/.83/.90. Of particular interest, however, is the gap between human and machine. On the original BSDS annotations, HED achieves ODS of .79 while human F score is .81, leaving a gap of just .02. On our annotations, however, HED drops to .69 while human F score increases to .90. Thus, unlike the original annotations, our dataset leaves substantial room for improvement of the state-of-the-art.

\section{Metrics and Baselines}\label{sec:evaluation}

We aim to develop measures to quantify algorithm performance on our data. We begin by reiterating that our rich annotations subsume many classic grouping tasks, including modal segmentation, edge detection, and figure-ground edge labeling. Indeed, our COCO dataset (5000 images) is an order of magnitude larger than BSDS (500 images), the previous de-facto dataset for these tasks. We encourage researchers to use our data to study these classic tasks; for well-established metrics we refer readers to~\cite{Arbelaez2011PAMI}.

Here we propose two simple metrics that focus on the most salient aspect of our dataset: the amodal nature of the segmentations. Predicting amodal segments requires understanding object interaction and reasoning about occlusion. Specifically, we propose to evaluate: (1) amodal segment quality and (2) pairwise depth ordering between regions. We additionally define strong baselines for each task.

All experiments are on the 5000 COCO annotations, split into 2500/1250/1250 images for train/val/test, respectively. We evaluate on val and reserve the test images for use in a possible future challenge as is best practice on COCO.

\begin{table*}
\subfloat[amodal segmentation evaluation\label{table:eval-amodal}]{
\centering\footnotesize\renewcommand\arraystretch{1.05}
\addtolength{\tabcolsep}{-1.1mm}\hspace{-3mm}
\begin{tabular}{@{\hskip 0mm}l|cccc|cccc|cccc@{\hskip 0mm}}
  & \multicolumn{4}{c|}{all regions}
  & \multicolumn{4}{c|}{things only}
  & \multicolumn{4}{c}{stuff only}\\
  & AR & AR\tss{N} & AR\tss{P} & AR\tss{H}
  & AR & AR\tss{N} & AR\tss{P} & AR\tss{H}
  & AR & AR\tss{N} & AR\tss{P} & AR\tss{H}\\
\shline
  DeepMask~\cite{pinheiro2015learning}
  & .378 & .456 & .407 & .248 & .422 & .470 & .473 & .279 & .248 & .367 & .242 & .199\\
  SharpMask~\cite{pinheiro2016learning}
  & .396 &\bf{.493} & .428 & .242 & .448 & \bf{.510} & \bf{.501} & .275 & .246 & .384 & .243 & .187\\
  ExpandMask\tss{S}
  & .384 & .460 & .415 & .256 & .427 & .474 & .480 & .284 & .258 & .374 & .250 & .212\\
  AmodalMask\tss{S}
  & .395 & .457 & .424 & .289 & .435 & .468 & .487 & .316 & .282 & .388 & .268 & .246\\
  ExpandMask
  & .417 & .480 & .428 & .327 & .456 & .495 & .488 & .351 & .305 & .387 & .278 & .289\\
  AmodalMask
  & \bf{.434} & .470 & \bf{.460} & \bf{.364} & \bf{.458} & .479 & .498 & \bf{.376} & \bf{.366} & \bf{.414} & \bf{.365} & \bf{.346}\\
\end{tabular}}
\subfloat[depth ordering evaluation\label{table:eval-order}]{
\centering\footnotesize\renewcommand\arraystretch{.9}
\addtolength{\tabcolsep}{-1.5mm}\hspace{6mm}
\begin{tabular}{@{\hskip 0mm}l|ccccc@{\hskip 0mm}}
  & Sharp & Expand & Amodal & Ground & Ground\\
  & Mask  & Mask   & Mask   & Truth  & Truth\\
  train-recall  & 45\% & 56\% & 59\% & 50\% & 100\%\\
  test-recall   & 41\% & 51\% & 54\% & 100\% & 100\%\\
\shline
  area              & .696 & .703 & .719 & .715 & .715\\
  y-axis            & .711 & .708 & .706 & .702 & .702\\
  OrderNet\tss{B}   & .753 & .764 & .770 & .770 & .765\\
  OrderNet\tss{M}   & .786 & .785 & .791 & .810 & .817\\
  OrderNet\tss{M+I} & \bf{.793} & \bf{.802} &\bf{.814} & \bf{.869} & \bf{.883}\\
\end{tabular}}
\Caption{(a) Amodal segmentation quality on the COCO validation set for multiple baselines and under no, partial, and heavy occlusion (AR\tss{N}, AR\tss{P}, AR\tss{H}). (b) Accuracy of pairwise depth ordering baselines applied to various segmentations results. See text for details.}
\end{table*}

\subsection{Amodal Segment Quality}

\textbf{Metrics}: To evaluate amodal segments, we adopt a popular metric for object proposals: average recall (AR), proposed in \cite{Hosang2015proposals} and used in the COCO challenges. To compute AR, segment recall is computed at multiple IoU thresholds (0.5-0.95), then averaged. To extend to our setting, we simply measure the IoU against the \emph{amodal} masks. We measure AR for 1000 segments per image and also separately for things and stuff. Finally, we report AR for varying occlusion levels $q$: none ($q$=0), partial (0$<$$q$$\le$.25), and heavy ($q$$>$.25), comprising 39\%, 31\% and 30\% of the data.

\textbf{Baselines}: We use \emph{DeepMask}~\cite{pinheiro2015learning} and \emph{SharpMask}~\cite{pinheiro2016learning}, current state-of-the-art methods for \emph{modal} class-agnostic object segmentation, as our first baselines. Next, inspired by Ke et al.~\cite{li2016amodal} (which is not directly applicable to our setting), we propose a deep network we call \emph{ExpandMask}. ExpandMask takes an image patch and a modal mask generated by SharpMask as input and outputs an amodal mask. Finally, we train a network, which we call \emph{AmodalMask}, to directly predict amodal masks from image patches. ExpandMask and AmodalMask share an identical network architecture with SharpMask (except ExpandMask adds an extra input channel and uses a slightly larger input size). However, while AmodalMask is run convolutionally, ExpandMask is evaluated on top of SharpMask segments.

We use the DeepMask and SharpMask publicly available code and pre-trained models. We implement ExpandMask and AmodalMask on top of the same codebase. Our models are initialized from the SharpMask network trained on the original modal COCO data. We finetune using our amodal training set. We also attempted to finetune our models using synthetic amodal data (\emph{ExpandMask\tss{S}} and \emph{AmodalMask\tss{S}}) by randomly overlaying objects masks from the original COCO dataset. For reproducibility, and to elucidate design and network choices, all source code will be released.

\begin{figure}\centering
\newcommand{\incp}[1]{\includegraphics[height=.17\linewidth]{figures/masks/foo_#1-fs8.png}}
\newcommand{\incr}[1]{\incp{#1_obj1} & \incp{#1_obj2} & \incp{#1_obj3} & \incp{#1_obj4}\\}
\renewcommand\arraystretch{.25}
\addtolength{\tabcolsep}{-1.9mm}
\begin{tabular}{cccc}
\incr{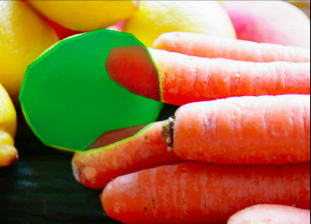} \incr{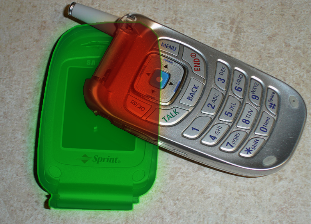} \incr{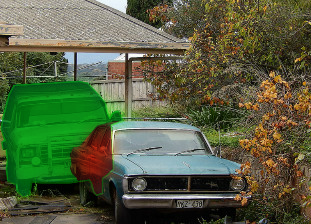}  \incr{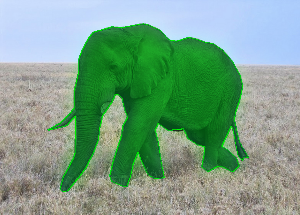}
GroundTruth & SharpMask & ExpandMask & AmodalMask\\
\end{tabular}
\Caption{Examples of amodal mask prediction (red indicates occlusion). SharpMask predicts \emph{modal} masks; ExpandMask and AmodalMask predict \emph{amodal} masks. The last row shows an unoccluded object, for which ExpandMask is overzealous.}\label{fig:qualitative-amodal}
\end{figure}

\textbf{Results}: AR for all methods is given in Table \ref{table:eval-amodal} and qualitative results are shown in Figure \ref{fig:qualitative-amodal}. SharpMask is a strong baseline, especially for things and under limited occlusion, which is its training setup. With more occlusion, the amodal baselines are superior, indicating these models can predict amodal masks (however, they are worse on unoccluded objects).  Using synthetic data improved AR on occluded regions over SharpMask but lagged the accuracy of using real training data. Finally, we note that human accuracy on this task is still substantially higher (see \S\ref{sec:consistency}).

\subsection{Pairwise Depth Ordering}

\textbf{Metrics}: Understanding full scene structure is challenging. Instead, we focus on evaluating pairwise depth ordering, which still requires reasoning about object interactions and spatial layout. Specifically, we report the accuracy of predicting which of two overlapping masks is in front. There are 36k/23k overlapping masks in the train/val sets.

Note that we have decoupled depth ordering from mask prediction. Since higher quality masks should be easier to order, we test each ordering algorithm with masks from multiple segmentation approaches. Specifically, for each ground truth mask we first find the best matching mask generated by a segmenter (with IoU of at least $0.5$), we then evaluate the depth ordering only on these matched masks.

\textbf{Baselines}: We start with two trivial baselines: order by area (smaller mask in front) and order by y-axis (mask closest to top in back). Next, we implemented a number of deep nets for this binary prediction task: OrderNet\tss{B} which takes two bounding boxes as input, OrderNet\tss{M} which takes two masks as input, and OrderNet\tss{M+I} which takes two masks and an image patch. OrderNet\tss{B} uses a 3 layer MLP while the other variants use pre-trained ResNet50 models~\cite{He2016} (modified slightly to account for varying number of input channels). We train and test a separate OrderNet model for each set of masks. For each prediction we run inference twice (with input order reversed) and average the results.

\textbf{Results}: We report results in Table \ref{table:eval-order}. In addition to ordering masks from multiple segmentation algorithms, we also train and test OrderNet on ground truth masks (with varying amount of training data) to capture the role of mask quality and data quantity on ordering accuracy. The naive heuristics (area and y-axis) both achieve about 70\% accuracy. OrderNet performs much better, with OrderNet\tss{M+I} achieving \app80\% accuracy on generated masks and \app90\% on ground truth. OrderNet benefits from better masks (performance increases in each row moving from left to right), and the percent of recalled pairs also affects results slightly (as there is more data for training). Considering the simplicity of our approach, these results are surprisingly strong.

\section{Discussion}\label{sec:discuss}

We presented a new dataset to study perceptual grouping tasks. The most distinctive feature of our dataset is that regions are annotated amodally: both the visible and occluded portions of regions are marked. The motivation is to encourage amodal perception, and reasoning about object interactions and scene structure. Extensive analysis shows that semantic amodal segmentation is a well-posed annotation task. We also provided evaluation metrics and strong baselines for the proposed tasks. We hope our dataset will help stimulate new research directions for the community.

\section*{Acknowledgements}
{\small We would like to thank Saining Xie and Yin Li for help with training the HED detector and to Lubomir Bourdev and Manohar Paluri and many others for valuable discussions and feedback.}

{\small\bibliographystyle{ieee}\bibliography{Amodal}}

\appendix
\section{Appendix: Annotation Details}

\subsection{Annotation Tool}

For our task we adopt the Open Surfaces \cite{bell2013opensurfaces} annotation tool developed by Bell \etal for material segmentation. The original tool allows for labeling multiple regions in an image by specifying a closed polygon for each region. The same tool was also adopted for annotation of COCO~\cite{mscoco2015}. The interface is simple and intuitive.

We extend the tool in a number of ways to support semantic amodal segmentation and facilitate annotation (see Figure~\ref{fig:screenshot}). We have added the following features:

\textit{Depth ordering:} An ordered list next to the image indicates the segment depth order. Annotators can rearrange the order by dragging items up and down in this list (see Figure \ref{fig:screenshot}). Moreover, visual feedback is given about depth order through the region fill overlaid on the image, allowing annotators to quickly determine the correct order, see Fig.~\ref{fig:depth+sharing}a.

\textit{Semantic annotation:} The same list used for specifying depth ordering is also used for naming each segment. The annotators enter free-form text for the segment names. All segments must be named for an annotation to be complete.

\textit{Edge sharing:} We extended polygon annotation to allow for `snapping' of a new polygon vertex to the closest existing polygon edge or vertex. This mechanism allows for easily annotating shared edges, see Figure \ref{fig:depth+sharing}b.

\textit{Polygon editing:} Finally, we add control for adding and removing vertices while editing existing polygons.

We will release the code for the modified annotation tool.

\begin{figure}\centering
\includegraphics[width=.48\textwidth]{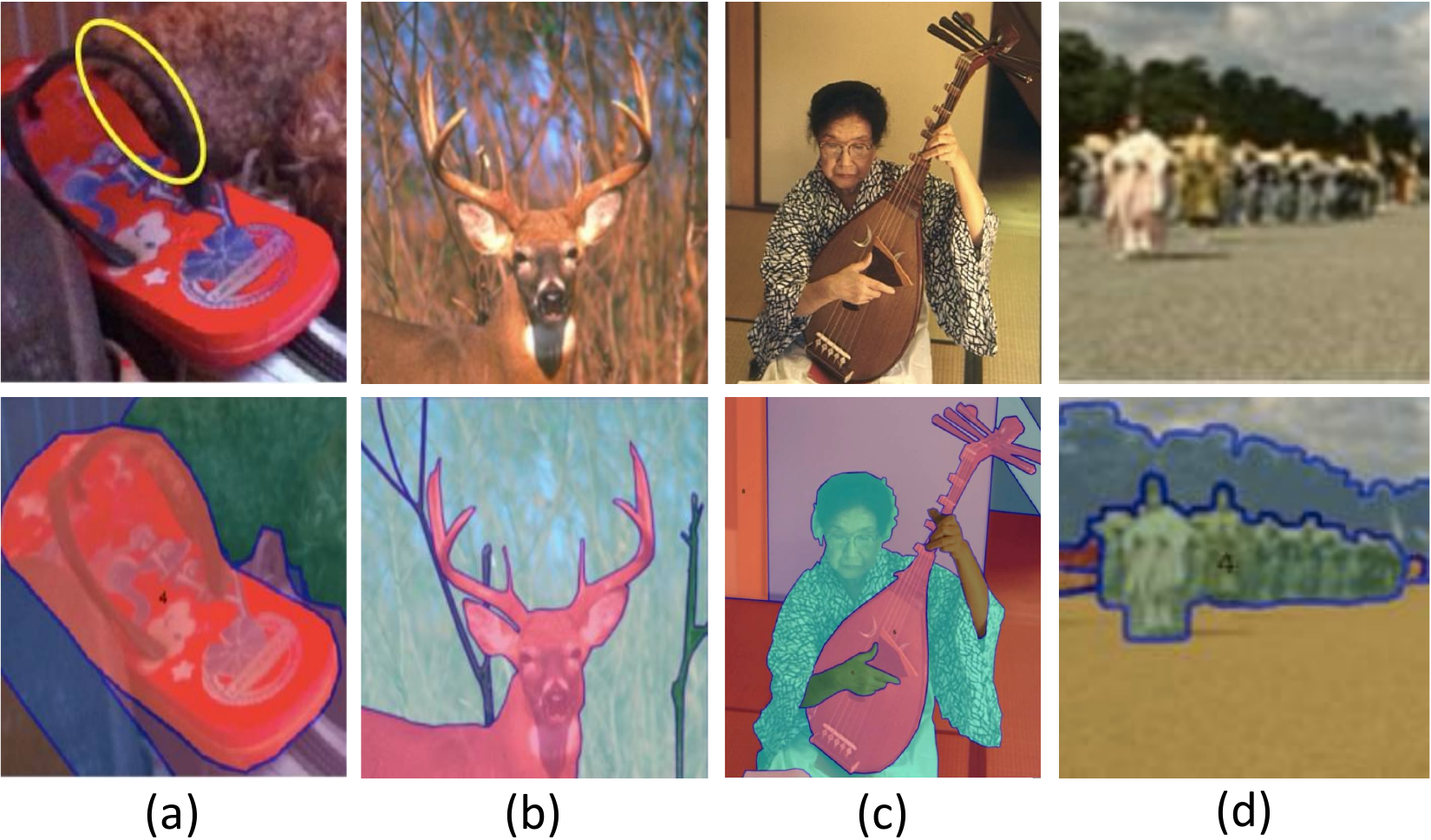}
\Caption{A few corner cases in annotation: (a) Annotators only label exterior boundaries, leaving holes as part of the region. (b) Annotators only label the most salient objects in blurry and cluttered backgrounds. (c) For regions with intertwined depth ordering, annotators are instructed to pick the depth ordering which is `least wrong' or to annotate object parts. (d) Annotators can mark a group of similar objects using a single segment.}
\label{fig:corner}
\end{figure}

\subsection{Corner Cases}

Although our annotation instructions are sufficient for most images, the following cases require special treatment:

\textit{Regions with holes}: We only annotate the exterior region boundaries, therefore each region is represented by a single segment. Holes are ignored (Figure~\ref{fig:corner}a).

\textit{Background objects}: For blurry objects in the background, annotators are asked to label only the most salient objects individually, rather than every detail (Figure~\ref{fig:corner}b).

\textit{Intertwined depth}: Two regions might not have a valid depth ordering (e.g., the woman holding the musical instrument in Figure~\ref{fig:corner}c). In such cases we instruct the annotators to pick the depth ordering which is `least wrong'. In extreme cases, annotators may label parts of an object so that visibility and occlusion information are correctly specified (e.g., by marking the woman's hands in Figure~\ref{fig:corner}c).

\textit{Groups}: For groups of similar objects (e.g. a crowd of people or bunch of bananas), annotators are instructed to mark a single region enclosing the entire group (Figure~\ref{fig:corner}d). Note that groups are often perceived as a single visual entity, so this form of annotation is quite natural.

\textit{Truncation}: Segments must be fully contained within the image boundaries, \ie regions extending beyond the image are \emph{not} annotated amodally (annotation outside the image is particularly challenging as the occluder is not visible).

\subsection{Annotators}

Rather than rely on a crowdsourcing platform, we utilize a pool of expert workers to perform all annotations. This allows us to specify more complex instructions than is typically possible with crowdsourcing platforms and iterate with workers until annotations reach a sufficient quality. We note, however, that if necessary we could move our annotation onto a crowdsourcing platform. This would require splitting a single image annotation into multiple separate and possibly redundant tasks, similarly to how annotation was performed on COCO~\cite{mscoco2015}.

While every image in BSDS is annotated by multiple workers, we also monitor individual worker quality. We differentiate between \emph{obvious errors}, which we ask workers to correct, and \emph{subjective judgments}, which differ between individuals and for which a clear criterion is harder to define. Each image annotation is manually checked, and obvious errors are sent back to the annotators for improvement. Subjective judgements, on the other hand, are left to annotators' discretion. Checking annotations for errors is a quick and lightweight process (and can also be crowdsourced).

Common obvious errors include incorrect depth ordering, missing foreground objects, regions annotated modally, and low quality polygons. These errors all explicitly violate the annotation instructions and are easily identifiable. On the other hand, common subjective judgements include the semantic label used, the exact location of hidden edges, and whether a region was sufficiently salient to warrant annotation. As mentioned, annotators are asked to correct obvious errors but not subjective judgements.

\begin{figure*}\centering
\newcommand{\incp}[1]{\includegraphics[width=.18\linewidth]{figures/edges/#1.png}}
\newcommand{\incr}[1]{\incp{#1_rgb} & \incp{#1_bsds_orig} & \incp{#1_bsds_ours} &\incp{#1_bsds_ours1} & \incp{#1_coco}\\}
\renewcommand\arraystretch{.5}
\addtolength{\tabcolsep}{-1.5mm}
\begin{tabular}{ccccc}
\incr{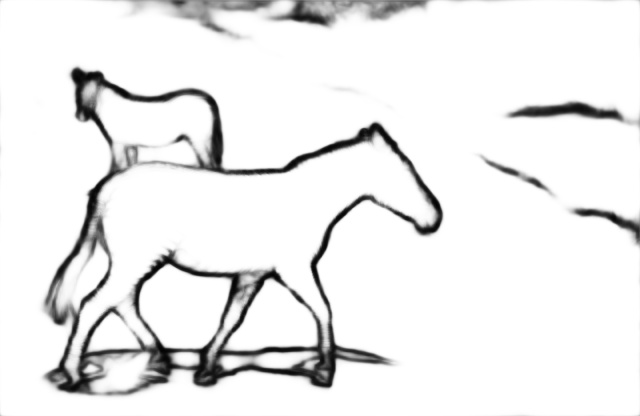}
\incr{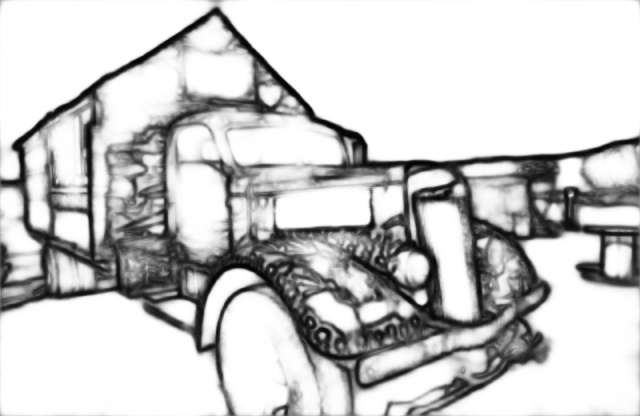}
\incr{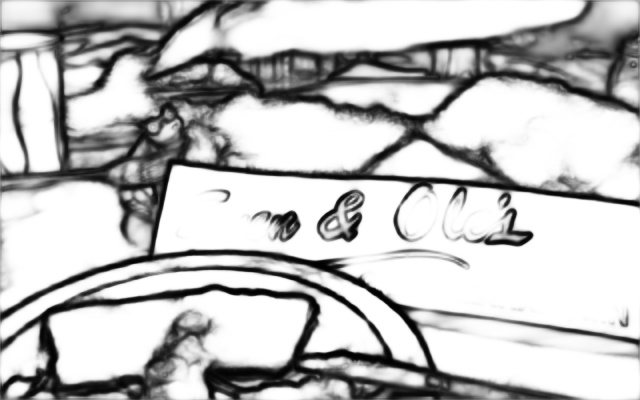}
\incr{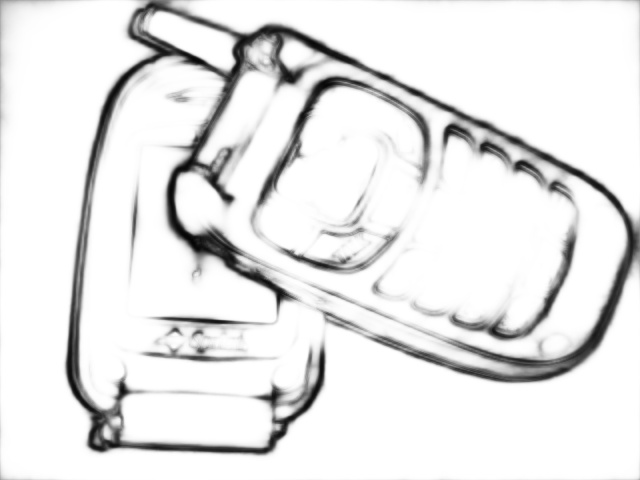}
(a) Image & (b) BSDS [original] & (c) BSDS-5 [ours] &(d) BSDS-1 [ours] & (e) COCO\\
\end{tabular}
\Caption{Edge detections for HED learned with different \emph{training} sets. (b) Using the original BSDS annotations results in dense edge maps with interior edges being detected. (c,d) Training with our BSDS edges (with either 1 or 5 annotators per image) results in sparser, more semantically meaningful edges. (e) Finally, training with our COCO edges yields qualitatively similar albeit slightly better results.}
\label{fig:qualitative-edges}
\end{figure*}

\section{Appendix: Edge Detection on COCO}

\begin{table}
\centering\small\renewcommand\arraystretch{1}
\renewcommand{\tabcolsep}{1.2mm}
\begin{tabular}{c|ccc|ccc}
 &\multicolumn{3}{c|}{\textbf{SE}~\cite{dollar2015fast}} & \multicolumn{3}{c}{\textbf{HED}~\cite{xie2015holistically}}\\
  train / test & bsds-5 & bsds-1 & coco-1 & bsds-5 & bsds-1 & coco-1 \\
\shline
bsds-5 & .630 & .543 & .522 & .694 & .615 & .583\\
bsds-1 & .628 & .540 & .520 & .690 & .609 & .575\\
coco-1 & .622 & .536 & .524 & .686 & .607 & .609\\
\end{tabular}
\Caption{Edge detection accuracy (ODS) versus the \emph{number of annotators per image.} Each row shows a different train setup and each column a different test setup. The number of annotators per image heavily affects test accuracy, but it makes little difference for training. Finally, switching the training set from BSDS to COCO has only a minor effect on SE but impacts HED more.}
\label{table:coco-vs-bsds}
\end{table}

To allow for the study of edge detectors on COCO, in this appendix we report the performance of the structured edges (SE)~\cite{dollar2015fast} and the holistically-nested edge detector (HED)~\cite{xie2015holistically} on COCO. Results of these detectors on the BSDS dataset~\cite{Arbelaez2011PAMI} (for both the original annotations and our annotations) were presented in \S\ref{sec:edges}. Here we train these state-of-the-art edge detectors on the 2500 COCO train images and test them on the 1250 image COCO val set.

We begin by noting that edge detection metrics~\cite{Arbelaez2011PAMI} are heavily impacted by the \emph{number of annotators per image}. The ground truth edges used for evaluation are the union of the human annotations and using more annotators per image results in denser edges for testing. In Table \ref{table:coco-vs-bsds}, we report edge detection accuracy versus the number of annotators per image using our annotations. During \emph{testing}, reducing the number of annotators per image lowers ODS substantially (even though the evaluated models are identical). On the other hand, reducing the number of annotations per image during \emph{training} leaves results largely unchanged.

\begin{table}
\centering\small\renewcommand\arraystretch{1.05}
\begin{tabular}{c|ccc}
 & ODS & AP & R50 \\
\shline
 SE~\cite{dollar2015fast}       & .524 & .474 & .519\\
 HED~\cite{xie2015holistically} & .609 & .493 & .741\\
\end{tabular}
\Caption{Edge evaluation for SE and HED on the COCO val set.}
\label{table:coco-edges}
\end{table}

From Table \ref{table:coco-vs-bsds} we also observe that results between COCO and BSDS are quite similar once the number of annotators per image is accounted for. We thus emphasize that while the edge detection accuracy on COCO appears to be worse than on BSDS (both using our annotations), this is an artifact of how accuracy is measured. We also note that while COCO only has one annotator per image, it has 10$\times$ more images than BSDS (5000 versus 500). Thus, more data-hungry approaches should benefit from COCO.

In Table \ref{table:coco-edges}, we report complete SE and HED edge detection results on the COCO validation set (training performed on the COCO train set). Our dataset provides a substantial challenge for current state-of-the-art edge detectors. Finally, in Figure \ref{fig:qualitative-edges}, we show qualitative HED edge detection results using different options for the training data.

\end{document}